\documentclass[hidelinks,onefignum,onetabnum]{siamart250211}

\usepackage{lipsum}
\usepackage{amsfonts}
\usepackage{graphicx}
\usepackage{epstopdf}
\usepackage{algorithmic}
\usepackage{mathrsfs}
\usepackage{rotating}
\usepackage{booktabs}

\usepackage{multirow}%
\usepackage{amsmath,amssymb}%
\usepackage{mathrsfs}%
\usepackage{xcolor}%
\usepackage{textcomp}%
\usepackage{manyfoot}%
\usepackage{booktabs}%
\usepackage{listings}%
\usepackage{array}

\ifpdf
  \DeclareGraphicsExtensions{.eps,.pdf,.png,.jpg}
\else
  \DeclareGraphicsExtensions{.eps}
\fi

\headers{Curvature-based geometric data analysis \& learning}{Y. Yadav, \& K. Xia}

\title{A roadmap for curvature-based geometric data analysis and learning
\thanks{This work is supported by Singapore Ministry of Education Academic Research fund Tier 1 grant RG16/23 and Tier 2 grants  MOE-T2EP20221-0003, MOE-T2EP50223-0014.}}

\author{
  Yasharth Yadav\thanks{School of Physical and Mathematical Sciences, 
  Nanyang Technological University, 637371 Singapore 
  (\email{yasharth001@e.ntu.edu.sg}, \email{xiakelin@ntu.edu.sg}).}
  \and
  Kelin Xia\footnotemark[2]
}

\usepackage{amsopn}

\ifpdf
\hypersetup{
  pdftitle={A roadmap for curvature-based geometric data analysis and learning},
  pdfauthor={Y. Yadav and K. Xia}
}
\fi

\begin{document}

\maketitle

\begin{abstract}
Geometric data analysis and learning has emerged as a distinct and rapidly developing research area, increasingly recognized for its effectiveness across diverse applications. 
At the heart of this field lies curvature, a powerful and interpretable concept that captures intrinsic geometric structure and underpins numerous tasks, from community detection to geometric deep learning.
A wide range of discrete curvature models have been proposed for various data representations, including graphs, simplicial complexes, cubical complexes, and point clouds sampled from manifolds. 
These models not only provide efficient characterizations of data geometry but also constitute essential components in geometric learning frameworks. 
In this paper, we present the first comprehensive review of existing discrete curvature models, covering their mathematical foundations, computational formulations, and practical applications in data analysis and learning. 
In particular, we discuss discrete curvature from both Riemannian and metric geometry perspectives and propose a systematic pipeline for curvature-driven data analysis. 
We further examine the corresponding computational algorithms across different data representations, offering detailed comparisons and insights.
Finally, we review state-of-the-art applications of curvature in both supervised and unsupervised learning.
This survey provides a conceptual and practical roadmap for researchers to gain a better understanding of discrete curvature as a fundamental tool for geometric understanding and learning.
\end{abstract}



\section{Introduction}\label{sec1}

Recent advances in machine learning have led to powerful methods for tackling a broad range of problems spanning domains such as computer vision \cite{simonyan2014very}, natural language processing \cite{otter2020survey, minaee2024large}, and speech recognition \cite{nassif2019speech}.
Most of these methods are applicable on data with an underlying Euclidean structure, where data points are represented as vectors within a high-dimensional Euclidean space. However, modern applications increasingly rely on data that exhibits a non-Euclidean structure.
Prominent examples include social networks in computational social science \cite{lazer2009computational}, sensor networks in wireless systems \cite{tubaishat2003sensor}, functional connectomes derived from brain imaging \cite{preti2017dynamic}, and gene regulatory networks in systems biology \cite{davidson2002genomic}.
Geometric deep learning aims to construct learning frameworks on non-Euclidean datasets \cite{bronstein2017geometric, cao2022geometric, weber2025geometric}.

At its core, geometric deep learning centers on incorporating the structure of data directly into model architectures.
One key development in geometric deep learning is the design of graph convolutional networks (GCNs) \cite{kipf2016semi}, which generalize convolutional neural networks (CNNs) to graph-structured data.
CNNs have achieved strong success in image processing \cite{lecun2010convolutional} due to their ability to leverage translation invariance as an inductive bias, enabling them to detect patterns in images regardless of spatial position.
However, in non-Euclidean domains such as graphs, there is no natural notion of translation.
Therefore, the standard convolution kernel fails to capture the same structural information. GCNs address this limitation by aggregating information from the local neighborhood of a node, effectively capturing both node features and local connectivity to learn useful representations.
This formulation has led GCNs to achieve state-of-the-art results across a wide range of graph learning tasks \cite{zhang2019graph}.
Other approaches to generalize CNNs to graphs include the use of recurrent neural networks combined with random walks \cite{gori2005new, sukhbaatar2016learning}, as well as convolutions defined in the spectral domain \cite{bruna2013spectral}.
A central theme across these models is their dependence on the structure and connectivity of data.
As such models gain prominence, it becomes essential to develop tools that can formally characterize this connectivity.
Discrete notions of curvature provide scalable and interpretable methods for quantifying the geometry of diverse data representations, including graphs, simplicial complexes, cubical complexes, and point clouds sampled on manifolds.

In Riemannian geometry, curvature is a local invariant of a manifold that measures its deviation from being Euclidean \cite{villani_optimal_2009, najman_modern_2017, jost2008riemannian}.
The notion of curvature has also been extended beyond smooth manifolds to more general metric spaces, often with a discrete structure \cite{stone1976combinatorial, lott2009ricci, bonciocat2009mass}.
A key qualitative feature shared by many of these discrete curvature models is their ability to distinguish bottleneck regions from highly clustered regions.
Nevertheless, each definition is fundamentally grounded in a distinct mathematical principle, reflecting a specific property of classical Riemannian curvature.
Among the most prominent definitions are Forman's discretization based on a combinatorial analog of the classical Bochner Weitzenböck formula in differential geometry \cite{forman_bochners_2003}, Ollivier's coarse Ricci curvature for metric spaces equipped with a measure or random walk \cite{ollivier_ricci_2007, ollivier_ricci_2009}, and Bakry-\'Emery's generalized lower bound of Ricci curvature \cite{bakry2006diffusions}.
Other definitions include Menger \cite{menger1930untersuchungen} and Haantjes \cite{haantjes1947distance} curvatures which capture geometric properties of triangles and paths in metric spaces respectively, curvature based on effective resistance inspired by electrical circuit theory \cite{devriendt2022discrete}, and a notion of sectional curvature that provides a global perspective on geometry \cite{joharinad_topology_2019, joharinad_geometry_2022}.
While many of these definitions have been studied theoretically \cite{bauer2013generalized, bauer_ollivierricci_2012, jost_olliviers_2014}, often in isolation, a few have also been evaluated empirically, primarily in the context of network data \cite{sreejith_forman_2016, sreejith_systematic_2017, samal_comparative_2018, mondal2024bakry}.

These notions of discrete curvature have also found several applications in machine learning. One of the earliest and most influential applications involves unsupervised node clustering or community detection in graphs, where two main strategies have been proposed, namely incremental removal of negatively curved edges \cite{sia2019ollivier, fesser2024augmentations} and updating edge weights using an associated discrete Ricci flow \cite{ni2019community, lai2022normalized}.
More recently, discrete curvatures have gained prominence in deep learning architectures, particularly in graph neural networks (GNNs) \cite{scarselli2008graph, defferrard2016convolutional}.
Notably, discrete curvatures have been used to characterize two central challenges in training graph neural networks, namely oversmoothing and oversquashing \cite{li2018deeper, alon2020bottleneck}.
Several curvature-informed graph rewiring methods have also been proposed as preprocessing steps that modify the graph structure to improve learning performance \cite{topping2021understanding, nguyen2023revisiting, fesser2024mitigating}.

The growing number of discrete curvature models and their expanding use in machine learning have resulted in a diverse but disconnected body of work that can be difficult to navigate for a newcomer.
This paper provides a structured overview of key curvature models, their computation across different data representations, and their state-of-the-art applications in machine learning.
Our goal is to support researchers and practitioners in understanding how curvature characterizes data geometry and how it can inform the design of more effective learning algorithms.

The current paper is structured as follows.
Section~\ref{sec2} presents widely used generalizations of curvature to general metric spaces, with particular emphasis on extensions of sectional, Ricci, and scalar curvature, and their theoretical motivations in Riemannian and metric geometry.
Building on this foundation, Section~\ref{sec3} presents a three-step pipeline for curvature-based data analysis. Specifically, it first introduces various topological representations of data, then presents several notions of discrete curvature and explains their computation through illustrative examples, and finally discusses how different curvatures can be used to extract meaningful geometric features from data at multiple scales.
Section~\ref{sec4} reviews state-of-the-art applications of discrete curvature in machine learning, spanning both supervised and unsupervised domains, with a particular focus on community detection, manifold learning, graph neural networks, and graph representation learning.

\section{Mathematical Background of Curvatures: Motivation from Riemannian and Metric Geometry}\label{sec2}

In Riemannian geometry, curvature is a local invariant of a metric space that measures its deviation from being Euclidean \cite{villani_optimal_2009}.
It is most commonly expressed in the form of \textit{Riemann curvature tensor} which captures the non-commutativity of the covariant derivative on a manifold \cite{najman_modern_2017}.
Several simplifications of the Riemann curvature tensor exist, namely \textit{sectional curvature}, \textit{Ricci curvature} and \textit{scalar curvature}.
Sectional curvature is a real-valued function defined on the set of all tangent planes in a manifold, and contains the same information as the curvature tensor. In other words, sectional curvature completely determines the curvature tensor.
Ricci curvature is a contraction of the curvature tensor and can be assigned to the vectors on a manifold. It is defined as the average of the sectional curvature of the tangent planes containing a given vector.
Scalar curvature is a further contraction of the curvature tensor, and is defined as the average of the Ricci curvatures of the tangent vector at a given point.
For a formal treatment of curvature on Riemannian manifolds, and existing works detailing the definitions and relationships between sectional, Ricci, and scalar curvatures, readers are referred to \cite{kobayashi1996foundations, lee2006riemannian, jost2008riemannian, najman_modern_2017}.

Most real-world datasets are often modeled by discrete spaces such as graphs or simplicial complexes (more details on topological representations of data in Section~\ref{sec4}, which lack a suitable differentiable structure.
Hence, it is not possible to directly extend the notion of curvature as defined in the Riemannian setting.
Instead, one can define geometric and topological properties of curvature bounds in Riemannian manifolds  -- such as convergence and divergence of geodesics, volume growth of distance balls, transportation distance between balls, eigenvalue bounds of the Laplace operator or control of harmonic functions --  and then study these properties in discrete metric spaces to construct appropriate analogs of curvature.
Over the past two decades, several notions of \textit{generalized curvature} for metric spaces have been proposed, each capturing a specific property of the Riemannian curvature tensor.
In this section, we review three widely studied generalizations of Ricci curvature—\textit{Forman–Ricci}, \textit{Ollivier–Ricci}, and \textit{Bakry–\'Emery–Ricci} curvature. 
These notions have been extensively adapted for discrete metric spaces, particularly graphs, and are motivated by fundamental principles from Riemannian geometry that guide the extension of curvature to combinatorial settings.
We further discuss a notion of sectional curvature formulated for general metric spaces based on intersection patterns of closed balls.
Finally, we review a generalization of scalar curvature for finite metric spaces that captures the volume growth aspect of Riemannian curvature.

\subsection{Ricci Curvature}

\subsubsection{Forman Ricci Curvature}

Forman's definition of Ricci curvature \cite{forman_bochners_2003} is based on a combinatorial analog of the classical  Bochner-Weitzenbo\"ck formula in differential geometry, which provides a relationship between curvature and the classical Riemannian Laplace operator \cite{berger2003panoramic, najman_modern_2017, weber2017characterizing}.
While this definition was originally introduced for \textit{weighted CW cell complexes} \cite{stallings1967lectures}, it is also applicable to polyhedral complexes, simplicial complexes and graphs, which are all special cases of cell complexes \cite{forman_bochners_2003, weber2016can}.
In Riemannian geometry, the Bochner-Weitzenbo\"ck formula provides a decomposition of the Riemannian Laplace operator as follows:
$$
\square_p = (\nabla_p)^* \nabla_p + F_p,
$$
Here, $\square_p$ is the Riemannian Laplace operator on $p$-forms on a compact Riemannian manifold $\mathcal{M}$, $\nabla_p$ is a covariant derivative operator, and $F_p$ is a 0th-order operator whose value at a point $x \in \mathcal{M}$ depends only on derivatives of the Riemannian metric at $x$.
Forman showed that it is possible to provide the following canonical decomposition in a discrete setting, which mimics the classical Bochner-Weitzenbo\"ck formula:
$$
\square_p = B_p + F_p,
$$
where, $\square_p$ is now the \textit{combinatorial Riemann-Laplace operator}, $B_p$ is the \textit{combinatorial Bochner (or rough) Laplacian}, and $F_p$ is the $p$th \textit{combinatorial curvature function}.
Further, for any $p$-cell $\alpha$, Forman defined the following function
$$
\mathcal{F}_p(\alpha) = \langle F_p(\alpha), \alpha \rangle,
$$
and noted that while $\mathcal{F}_p$ has no clear geometric interpretation for $p>1$ in the Riemannian setting, it is equal to the Ricci curvature when $p=1$. In other words, the \textit{Forman-Ricci curvature} for any 1-cell or edge $e$ is equal to $\mathcal{F}_1(e)$. This definition is quite general and is applicable to any weighted cell complex.

\subsubsection{Ollivier Ricci Curvature}

In Riemannian geometry, Ricci curvature governs the growth of volumes in the direction of a tangent vector at a given point on the manifold.
Building on this key property of classical Ricci curvature, Ollivier introduced the concept of \textit{coarse Ricci curvature} for metric spaces equipped with a measure or random walk \cite{ollivier_ricci_2007, ollivier_ricci_2009}.
This notion has been extensively explored in discrete metric spaces, particularly in graphs \cite{lin_ricci_2010, lin_ricci_2011, bauer_ollivierricci_2012, jost_olliviers_2014, samal_comparative_2018}.
Moreover, this is the only adaptation of Ricci curvature to graphs that is known to converge to the Riemannian version of Ricci curvature \cite{sturm2006geometry, lott2009ricci, van2021ollivier}.

Ollivier's definition is based on the following observation about Ricci curvature on Riemannian manifolds.
In regions of positive Ricci curvature, the centers of balls are closer than the balls themselves, whereas in regions of negative Ricci curvature, the balls are farther apart than their centers.
Here, if one thinks of a ball as measure of mass 1 around some point, then one can define a notion of distance between balls using the well-known \textit{transportation distance} or \textit{Wasserstein distance} \cite{vaserstein1969markov}.
The distance between the centers of the balls can be determined solely based on the metric of the underlying space.
Formally, let $m_x$ and $m_y$ be measures around two close enough points $x$ and $y$ respectively.
The distance between $x$ and $y$ is denoted as $d(x,y)$ and the Wasserstein distance between $m_x$ and $m_y$ is denoted as $W_1(m_x, m_y)$.
The Ollvier-Ricci curvature along $xy$ is given by \cite{ollivier2011visual}:
$$
\kappa(x, y) = 1 - \frac{W_1(m_x, m_y)}{d(x, y)}.
$$
The above definition reduces to the definition of classical Ricci curvature on Riemannian manifolds.
More precisely, if $v$ is the the unit tangent vector at $x$ directed along the geodesic $\overline{xy}$, then
$$
\kappa(x, y) = \frac{\varepsilon^2}{2(n + 2)} \text{Ric}(v) + O(\varepsilon^3 + \varepsilon^2 d(x, y)).
$$

\subsubsection{Bakry and \'Emery Ricci Curvature}

Similar to Forman's notion of Ricci curvature, Bakry and \'Emery proposed a version of Ricci curvature for general metric spaces, rooted in the Bochner-Weitzenböck decomposition \cite{bakry2006diffusions, najman_modern_2017}.
However, unlike Forman's approach which focuses on the decomposition of the Riemannian Laplace operator, this formulation is based on \textit{Bochner's formula} \cite{najman_modern_2017, cushing2020bakry, mondal2024bakry} providing a generalized lower bound for Ricci curvature and governing the Laplacian eigenvalues through these lower bounds.
It states that, for all smooth functions $f$ on a Riemannian manifold of dimension $n$,
$$
\frac{1}{2} \Delta |\nabla f|^2(x) = |\text{Hess} \, f|^2(x) + \langle \nabla \Delta f(x), \nabla f(x) \rangle + \text{Ric}(\nabla f(x)),
$$
where \(\text{Hess}\) denotes the Hessian, \(\text{Ric}\) denotes the Ricci tensor, whereas $\Delta$ and $\nabla$ denote the Laplacian and gradient respectively.
If the Ricci curvature is bounded from below by $K_x$ at $x \in \mathcal{M}$ then $\text{Ric}(v) \geq K_x |v|^2$ for all $ v \in T_x \mathcal{M}$. Further, using the inequality $|\text{Hess} \, f|^2(x) \geq \frac{1}{n}(\Delta f(x))^2$, we obtain the following curvature dimension inequality \cite{pouryahya2016bakry, najman_modern_2017, cushing2020bakry}:
$$
\frac{1}{2} \Delta |\nabla f|^2(x) - \langle \nabla \Delta f(x), \nabla f(x) \rangle \geq \frac{1}{n} (\Delta f(x))^2 + K_x |\nabla f(x)|^2.
$$
Bakry and \'Emery utilized bilinear operators $\Gamma$ and $\Gamma_2$ for smooth functions $f,h \in C^{\infty}(\mathcal{M})$, defined as
$$
\Gamma(f, h) := \frac{1}{2} \left[ \Delta(fh) - f \Delta h - h \Delta f \right], \text{ and}
$$
$$
\Gamma_2(f, h) := \frac{1}{2} \left[ \Delta \Gamma(f, h) - \Gamma(f, \Delta h) - \Gamma(h, \Delta f) \right].
$$
Further, denote $\Gamma(f) := \Gamma(f,f)$ and $\Gamma_2(f) := \Gamma_2(f,f)$.
It can be seen that $\Gamma(f,f) = |\nabla f|^2$ and $\Gamma_2(f, f) = \frac{1}{2} \Delta |\nabla f|^2 - \langle \nabla \Delta f(x), \nabla f(x) \rangle$. As a result, the curvature dimension inequality takes the following form,
$$
\Gamma_2(f)(x) \geq \frac{1}{n} (\Delta f(x))^2 + K_x \Gamma(f)(x).
$$

A key advantage of the $\Gamma$-calculus framework of Bakry and \'Emery is that the $\Gamma$ and $\Gamma_2$ operators are defined merely via the Laplacian.
This formulation enables a natural extension of the curvature dimension inequality to general metric spaces and allows one to derive a lower bound on Ricci curvature, provided the Laplacian is specified on such spaces.
Beyond their ability to provide lower bounds of Ricci curvature, these operators also share a deep connection with the geometry of the underlying manifold $\mathcal{M}$.
In particular, $\Gamma(f,h)$ is closely related to the \textit{carré du champ} formula, which expresses the Riemannian metric of vector fields $\nabla f$ and $\nabla h$ in terms of the Laplacian,
$$
g(\nabla f, \nabla h) = \frac{1}{2} \left[ f \Delta h + h \Delta f - \Delta(fh) \right] .
$$
This connection not only provides a clear geometric interpretation of the $\Gamma$ operators, but has also been utilized by recent works in \textit{diffusion geometry} to recover the Riemann curvature tensor from point clouds sampled on a manifold \cite{jones2024diffusion, jones2024manifold}.

\subsection{Sectional Curvature}

Joharinad and Jost \cite{joharinad_topology_2019, joharinad_geometry_2022} introduced a notion of sectional curvature to quantify the intersection patterns of closed balls in general metric spaces.
This notion of sectional curvature is inspired by Gromov's metric characterizations of non-positive curvature.
Unlike the classical definition of curvature which measures the deviation from Euclidean metric, this notion measures the deviation from tripod spaces.
In other words, tripod spaces emerge as a natural choice of ``model space'' for comparing different metric spaces.
A geodesic length space $(X, d)$ is known as a tripod space if for any three points $x_1, x_2, x_3 \in X$, there exists a point $m \in X$, also known as the $median$, such that
$$
d(x_i,m) + d(x_j,m) = d(x_i,x_j), \text{ for } 1 \leq i < j \leq 3.
$$
A tripod space $(X,d)$ satisfies the following important property which is based on intersection patterns of closed balls around three points $x_1, x_2, x_3 \in X$ that \textit{do not} lie on a straight line.

Given three positive real numbers $r_1, r_2, \text{ and } r_3$, such that $r_i + r_j \geq d(x_i, x_j)$, where $1 \leq i < j \leq 3$,
$$
\bigcap_{i=1,2,3} B(x_i, r_i) \neq \varnothing.
$$
Here, $B(x_i, r_i) := \{ x \in X : d(x_i, x) \leq r_i \}$ denotes a \textit{closed ball} centered at $x_i$ with radius $r_i$.
The aforementioned tripod property is crucial for defining sectional curvature of metric spaces.
The key idea is to measure the extent to which closed balls in metric spaces need to be enlarged in order to intersect.
In particular, for three points $x_1, x_2, x_3 \in X$, consider three closed balls $B(x_1, r_1)$, $B(x_2, r_2)$ and $B(x_3, r_3)$ respectively.
The radii of these closed balls are chosen such that they intersect pairwise at the boundary.
This condition can be expressed by the following set of equations:
$$
r_i + r_j = d(x_i, x_j), 1 \leq i < j \leq 3.
$$
The above system of equations has a unique solution which is given by \textit{Gromov products}, as below
\begin{align*}
r_1 = \frac{1}{2}[d(x_1,x_2) + d(x_1, x_3) - d(x_2, x_3)], \\
r_2 = \frac{1}{2}[d(x_1,x_2) + d(x_2, x_3) - d(x_1, x_3)], \\
r_3 = \frac{1}{2}[d(x_1,x_3) + d(x_2, x_3) - d(x_1, x_2)].
\end{align*}

Given the above initial construction, we proportionally increase the radii of the three balls until we get a non-empty intersection. In other words, we expand the radius of each of the three balls by an \textit{expansion constant} $\rho(x_1, x_2, x_3)$, which is defined as the greatest lower bound or the \textit{infimum} of numbers $\mu$, such that
$$
\bigcap_{i=1,2,3} B(x_i, \mu r_i) \neq \varnothing.
$$
The above expansion constant $\rho(x_1, x_2, x_3)$ is the \textit{sectional curvature} associated with the triple of points $x_1$, $x_2$ and $x_3$ in a general metric space $(X,d)$.
A more formal expression of sectional curvature is as follows.
$$
\rho(x_1, x_2, x_3) := \inf_{x \in X} \max_{i=1,2,3} \frac{d(x_i, x)}{r_i}
$$

For complete metric spaces $1 \leq \rho(x_1, x_2, x_3) \leq 2$.
The lower bound is achieved for any three points in a tripod space. In other words, for tripod spaces, a pairwise intersection of three closed balls guarantees a common intersection.
Further, if the closed balls intersect pairwise at their boundaries, the common intersection results in a point $m \in X$ which is the median.
As a result, sectional curvature is an invariant associated with triples of points in a metric space which measures the deviation from tripod property.
Larger values of sectional curvature indicate that a metric space is further away from being a tripod space.
For a detailed discussion of the above definition and its generalizations to more than three points, see Appendix~\ref{secA1}.

\subsection{Scalar Curvature}
Given a Riemannian manifold $\mathcal{M}$, the scalar curvature is a function $S: \mathcal{M} \to \mathbb{R}$ that assigns a  real number $S(x)$ to each point $x$ in the manifold.
While it is proportional to the notion of Gaussian curvature on surfaces, it can be extended to higher-dimensional manifolds as well. 
Building on this classical concept, Hickok and Blumberg  \cite{hickok2023intrinsic} proposed a generalization of scalar curvature that can be estimated on a finite sample $X \subset \mathcal{M}$. 
The set $X$ comprises a finite metric space generated by independent draws from a probability density function $\rho: \mathcal{M} \to \mathbb{R}_+$.
This generalization of scalar curvature captures the \textit{volume growth} property of the Riemannian curvature.

Formally, for a manifold $\mathcal{M}$ of dimension $n$, let $B^{\mathcal{M}}(x, r)$ denote a geodesic ball of radius $r$ centered at $x \in \mathcal{M}$, let $\operatorname{vol}(B^{\mathcal{M}}(x, r))$ denote its volume and $v_n r^n$ the volume of a Euclidean $n$-ball of radius $r$.
The relationship between the scalar curvature $S(x)$ at $x \in \mathcal{M}$ and the volume of $B^{\mathcal{M}}(x, r)$ (as $r \to 0$) follows the below equation:
$$
\frac{\operatorname{vol}(B^{\mathcal{M}}(x, r))}{v_n r^n}
= 1 - \frac{S(x)}{6(n + 2)}r^2 + O(r^4).
$$

The process of obtaining an estimate of the scalar curvature $S(x)$ at a point $x \in X$ involves computing the volumes of geodesic balls $B^{\mathcal{M}}(x, r)$ for a sequence of increasing radii $r$, and subsequently fitting a quadratic polynomial to the estimated ratios $\operatorname{vol}(B^{\mathcal{M}}(x, r)) / (v_n r^n)$. 
The volumes of geodesic balls are estimated directly from the \textit{distance matrix} $d_X$ of the finite metric space $X \subset \mathcal{M}$. 
If $\widehat{C}(x)$ denotes the quadratic coefficient of the fitted curve, the estimated scalar curvature is given by
$$
\widehat{S}(x) = -6(n + 2)\widehat{C}(x).
$$

The above estimate of scalar curvature depends only on the pairwise distances between points in the finite metric space $X$. 
This formulation conceptually resembles the notion of sectional curvature discussed in the preceding section, where curvature is defined based on the pairwise distances among triplets of points rather than from an explicit embedding of these points.
Since this scalar curvature estimate is fundamentally intrinsic in nature, it can be easily applied to general metric spaces, including discrete spaces such as graphs equipped with a shortest-path metric.
Thus, it provides a principled and mathematically rigorous generalization of scalar curvature to arbitrary finite metric spaces, naturally complementing existing generalizations of Ricci and sectional curvature.

Some previous works have also attempted to define scalar curvature in discrete metric spaces by averaging or summing Ricci curvature functions, such as Forman or Ollivier–Ricci curvature \cite{sandhu2016ricci, sreejith_forman_2016}.
Such aggregation-based definitions are inspired by the Riemannian formulation where scalar curvature arises as the trace (or contraction) of the Ricci curvature.
Nevertheless, most constructions remain heuristic and do not provide formal guarantees of convergence to scalar curvature in the Riemannian setting.
Recently, Hickok and Blumberg  \cite{hickok2025discrete} defined \textit{discrete scalar curvature} of a vertex in a graph as the weighted sum of Ollivier-Ricci curvatures of its incident edges. 
Importantly, their weighting scheme admits the following geometric interpretation.
When vertices correspond to points on a manifold and edge weights $w(x, y)$ represent their geodesic distances $d_{\mathcal{M}}(x, y)$, the formulation naturally assigns a higher weight on point pairs that are farther apart.
Furthermore, this weighting scheme yields a curvature function that provably converges to a constant multiple of the scalar curvature in the Riemannian setting.
An explicit expression of this discretization is presented in Section~\ref{sec3}.

Other attempts to generalize scalar curvature to finite metric spaces include estimates based on the Gauss–Codazzi equation and the second fundamental form \cite{sritharan2021computing}, approaches using persistent homology to infer curvature from sampled point clouds \cite{bubenik2020persistent}, methods for estimating Gaussian curvature on constant-curvature surfaces \cite{cazals2005estimating}, and formulations that define curvature through geometric intersections of Euclidean balls \cite{chazal2009stability}.

\section{Curvature-based Geometric Data Analysis}\label{sec3}


In the previous section, we reviewed widely studied notions of generalized curvature and their theoretical motivations from Riemannian and metric geometry.
In the rest of the paper, we transition from theory to practice by presenting a general framework for extracting geometric features from data using curvature.
In addition, we provide a comprehensive review of the applications of curvature in machine learning, with a focus on graph neural networks, commmunity detection and manifold learning. The pipeline for curvature-based geometric data analysis -- and more broadly, \textit{geometric data analysis} -- can be divided into three main steps.

The first step is to extract a suitable representation of the underlying data \cite{oudot2017persistence, otter_roadmap_2017, bick2023higher}.
Depending on the application domain, data can take several forms.
For example, relational data is typically available in the form of networks, digital images and volumetric data are made of pixels (in 2D) and voxels (in 3D) respectively, whereas other types of data can be construed as point clouds or finite metric spaces. The second step is to apply an appropriate definition of curvature on the extracted topological representation. Most representations of data, in particular graphs and simplicial complexes, are discrete and often admit a well-defined metric structure.
This allows one to carefully formulate a notion of \textit{discrete curvature} that is well-adapted to these representations.
Several notions of curvature, including those introduced in the previous section and others discussed later, can be rigorously defined on graphs.
Some, such as Forman-Ricci curvature, also extend naturally to higher-order objects like simplicial complexes. The third step is to extract meaningful geometric features after computing curvature on the data. Forman-Ricci and Ollivier-Ricci curvatures are edge-based geometric invariants and hence can naturally featurize the edges or connections in the data. Bakry-\'Emery curvature, on the other hand, can be used to featurize the vertices or individual data points. Forman-Ricci curvature can also featurize higher-order relations in data. 

\subsection{Data Representations}\label{sec3.1}

Various data representations can be considered depending on the type of underlying data.
Relational data can be represented as graphs, abstract simplicial complexes, clique complexes, or hypergraphs. Digital images are often represented as cubical complexes, whereas point cloud data can be described using Vietoris-Rips complexes and alpha complexes.
Here, we provide mathematical definitions for some of these data representations.
A more detailed discussion of these representations (and associated methods in topological data analysis, such as persistent homology) can be found in  \cite{zomorodian2004computing,carlsson2009topology, otter_roadmap_2017, munch_users_2017, edelsbrunner2022computational}.

\subsubsection*{\textit{Graph}}

An \textit{undirected graph} \( \mathcal{G} = (\mathcal{V}, \mathcal{E}) \) consists of
(i) a finite set of vertices \( \mathcal{V} = \{v_1, v_2, \dots, v_N\} \), and
(ii) a set of edges \( \mathcal{E} \subseteq \{ \{v_i, v_j\} \mid v_i, v_j \in \mathcal{V} \} \), which represent \textit{unordered} pairs of vertices in \( \mathcal{V} \).
A \textit{directed graph} \( \mathcal{G} = (\mathcal{V}, \mathcal{E}) \) includes (i) a finite set of vertices \( \mathcal{V} = \{v_1, \dots, v_N\} \), and
(ii) a set of directed edges \( \mathcal{E} \subseteq \mathcal{V} \times \mathcal{V} \), consisting of \textit{ordered} pairs of vertices.
A graph is called \textit{simple} if it has no self-loops, meaning a vertex cannot be connected to itself by an edge.
A graph \( \mathcal{G} \) with \( N \) vertices can be represented by an \( N \times N \) adjacency matrix \( \mathbf{A} \), where \( A_{ij} = 1 \) if vertices \( v_i \) and \( v_j \) form an edge, and \( A_{ij} = 0 \) otherwise. If the graph \( \mathcal{G} \) is undirected, its adjacency matrix is symmetric.

Graphs have a simple combinatorial structure, making them an interpretable and flexible tool for modeling pairwise relationships in datasets. In many cases, datasets are relational in nature, directly encoding connections such as collaborations, co-occurrences, or flows.
These relationships can be represented as edges in a graph \cite{benson2016higher, battiston2020networks}.
In other types of datasets, such as point clouds or finite metric spaces, each point is represented by a vertex, and edges are inferred in a manner robust to small perturbations, enabling the recovery of topological properties of the underlying space \cite{oudot2017persistence}.
An example of a graph constructed using such an approach is the $\epsilon$-neighborhood graph \cite{otter_roadmap_2017}.

In many applications, graphs with weighted edges are of significant interest \cite{barrat2004architecture, horvath2011weighted}.
A \textit{weighted graph} \( \mathcal{G} = (\mathcal{V}, \mathcal{E}) \) has edges \( \{v_i, v_j\} \in \mathcal{E} \) that have an associated weight \( w_{ij} \neq 0 \).
The concept naturally extends to directed graphs as well.
For any weighted graph \( \mathcal{G} \), the \textit{weighted adjacency matrix} \( A \) is defined as:
$A_{ij} = w_{ij} \quad \text{if} \quad (v_i, v_j) \in \mathcal{E}, \quad \text{and} \quad A_{ij} = 0 \quad \text{otherwise}$.

\subsubsection*{\textit{Abstract Simplicial Complex}}

Roughly, an abstract simplicial complex is a generalized version of a graph, formed by combining vertices, edges, triangles, tetrahedra, and their higher-dimensional analogues.
Formally, it is a collection \( \mathcal{K} \) of non-empty subsets of a set \( \mathcal{V} \), satisfying two key conditions.
First, if \( \sigma \in \mathcal{K} \) and \( \tau \subseteq \sigma \), then \( \tau \) must also belong to \( \mathcal{K} \).
Second, for every \( v \in \mathcal{V} \), the singleton \( \{v\} \) is included in \( \mathcal{K} \).
The elements of \( \mathcal{V} \) are called the vertices of \( \mathcal{K} \), and the elements of \( \mathcal{K} \) are referred to as simplices.
In the rest of the paper, we will use $v$ to refer to the singleton $\{ v \}$.

A simplex is said to have dimension \( p \) (or is a \( p \)-simplex) if it contains \( p+1 \) elements.
The set of all $p$-simplices is denoted by $\mathcal{K}_p$.
The dimension of a simplicial complex \( K \) is defined as the highest dimension among its simplices.
Graphs, consisting only of vertices and edges, can be interpreted as 1-dimensional simplicial complexes.
A \textit{face} of a \( p \)-simplex is any of its \( (p-1) \)-dimensional subsimplices formed by removing exactly one vertex.
The set of all faces of $\sigma \in \mathcal{K}_p$ is given by $\mathrm{Face}(\sigma) = \left\{ \tau \in \mathcal{K}_{p-1} \,\middle|\, \tau \subset \sigma \right\}.$
Conversely, a \textit{coface} of a \( k \)-simplex \( \sigma \in \mathcal{K}_p \) is any \( (p+1) \)-dimensional simplex \( \eta \in \mathcal{K}_{p+1} \) such that \( \sigma \subset \eta \).
The set of all cofaces of \( \sigma \) is given by $\mathrm{Coface}(\sigma) = \left\{ \eta \in \mathcal{K}_{p+1} \,\middle|\, \sigma \subset \eta \right\}$.

Abstract simplicial complexes have emerged as powerful combinatorial structures in the field of topological data analysis due to their ability to represent arbitrary topological spaces \cite{otter2017roadmap}.
Common input data that can be modeled with abstract simplicial complexes \cite{carlsson2009topology, edelsbrunner2008persistent, ghrist2014elementary, edelsbrunner2022computational} include point clouds in metric spaces and network data (nodes and relations).
We also provide definitions of two widely utilized abstract simplicial complexes: the Vietoris-Rips complex and the clique complex.

\subsubsection*{\textit{Vietoris-Rips Complex}}

Let \( X \) be a finite set of points in a \( d \)-dimensional Euclidean space \( \mathbb{R}^d \).
To define the Vietoris-Rips complex at scale \( \epsilon \), denoted as $\mathrm{Rip}_{\epsilon}(X) $, each point in \( X \) is associated with a closed ball of diameter \( \epsilon \) centered at that point.
\( \mathrm{Rip}_{\epsilon}(X) \) is constructed such that it contains a \( p \)-simplex if and only if the distance between any two points in this \( p \)-simplex is at most \( \epsilon \). Formally.
$$
\mathrm{Rip}_{\epsilon}(X) = \left\{ \sigma \subseteq X \ \middle| \ d(x, y) \leq \epsilon \ \text{for all} \ x, y \in \sigma \right\}
$$
The construction of a Vietoris-Rips complex depends on a predefined parameter \( \epsilon \).
Determining the appropriate value of \( \epsilon \) that accurately reflects the underlying topology of the data is a non-trivial task.
To address this, the concept of \textit{filtration} is introduced \cite{edelsbrunner2002topological, carlsson2009topology}, which involves progressively increasing the value of \( \epsilon \) to generate a sequence of topological spaces.
These spaces are represented as a nested family of simplicial complexes at multiple scales.
To summarize, Vietoris-Rips complex is a type of abstract simplicial complex defined on point clouds or finite metric spaces.
When combined with a filtration, the Vietoris-Rips complex provides a multiscale topological representation of data.

\subsubsection*{\textit{Clique Complex}}

A clique complex is defined on graphs and has a straightforward construction.
Given a graph \( \mathcal{G} \), a clique is a set of vertices that form a complete subgraph in \( \mathcal{G} \).
The clique complex of \( \mathcal{G} \) is formed by associating a \( p \)-simplex to a clique with \( p+1 \) vertices.
Note that a clique complex is also a type of abstract simplicial complex, as any subset of a clique is itself a clique.

Clique complexes provide a generalized topological framework for representing network data, where nodes represent entities and edges signify pairwise connections \cite{giusti2015clique, zomorodian2010tidy, burgio2021network}.
These complexes extend beyond pairwise relationships by capturing group interactions, commonly referred to as \textit{higher-order interactions} \cite{bick2023higher, battiston2020networks}, which involve three or more agents (or nodes) simultaneously.

\subsubsection*{\textit{Cubical Complex}}

A cubical complex is constructed from a union of vertices, edges, squares, cubes, and higher-dimensional hypercubes.  Formally, one proceeds by defining an elementary interval $I \subset \mathbb{R}$ as a closed interval of the form $I = [l, l+1]$ or $I = [l, l]$, for some $l \in \mathbb{Z}$.
An elementary cube $C$ is a finite product of elementary intervals, that is,  $C = I_1 \times I_2 \times \cdots \times I_d \subset \mathbb{R}^d$.
A set $\mathcal{Q} \subset \mathbb{R}^d$ is known as a cubical complex if it can be given by a finite union of elementary cubes.

Digital images have a natural cubical structure. 2D digital images are described as pixels whereas 3D images are described as voxels. Thus, cubical complexes provide a suitable topological representation of digital images.  One way to build a cubical complex from a digital image \cite{wagner2011efficient} is by assigning every pixel (respectively, voxel) to a vertex. Subsequently, one can connect vertices of adjacent pixels (respectively, voxels) by an edge and fill in the resulting squares (respectively, cubes). The weight of each vertex is the gray value of the pixel, and the weights of edges (respectively, squares) are the maximum values of the adjacent vertices (respectively, edges).

\subsubsection*{\textit{Hypergraph}}

Let $\mathcal{V} = \{1, \ldots, N\}$ be a finite set and let $\mathcal{E} \subseteq \mathscr{P}(\mathcal{V})$ be a finite collection of nonempty subsets of $\mathcal{V}$. Here, $\mathscr{P}(\mathcal{V})$ denotes the power set of $\mathcal{V}$, consisting of all possible subsets of $\mathcal{V}$.
The set $\mathcal{V}$ consists of vertices, the elements in the set $\mathcal{E}$ are known as hyperedges, and the tuple $\mathcal{H} = (\mathcal{V}, \mathcal{E})$  \cite{berge1985graphs} is called a hypergraph.

Hypergraphs are often used to model relational data where relationships extend beyond simple pairwise interactions.  For example, interactions between four vertices can be represented by a hyperedge of cardinality four. Like simplicial complexes, hypergraphs generalize graphs by capturing higher-order relationships. However, a key distinction between hypergraphs and simplicial complexes lies in the definition and completeness of boundary operators on their respective chain groups. Specifically, while boundary operators in simplicial complexes are well-defined, ensuring that the boundary of a simplex is composed of lower-dimensional simplices (which are also present in the simplicial complex), this property does not hold for hypergraphs. In other words, the boundary elements of a hyperedge may include subsets that are not themselves hyperedges, making the notion of a boundary operator ill-defined in general for hypergraphs.

\subsection{Discrete Curvature for Data Analysis}\label{sec3.2}

In Section~\ref{sec2}, we reviewed four generalizations of curvature.
We first examined two approaches that generalize classical Ricci curvature: the combinatorial approach of Forman and the optimal-transport approach of Ollivier.
Next we covered the $\Gamma$-calculus lower bound due to Bakry and \'Emery, and a metric generalization of sectional curvature.
In the present section we revisit these definitions in fully \textit{discrete} form, emphasizing their computation on topological representations generated from data.
In discrete settings, we will refer to these curvatures as \textit{Forman-Ricci curvature}, \textit{Ollivier-Ricci curvature}, \textit{Bakry-\'Emery curvature}, and \textit{sectional curvature}.
We then review three additional definitions of discrete curvature that have recently emerged in network analysis, namely \textit{Menger-Ricci curvature}, \textit{Haantjes-Ricci curvature} and \textit{resistance curvature}.

Table~\ref{Tab:AllCurvs} provides a summary of the mathematical notations used to express these seven curvature notions, their theoretical motivations, and key references for further reading.
The seven notions of discrete curvature can be categorized into three types based on the geometric object on which they are defined: scalar curvature $p(v)$ defined on vertices; Ricci curvature $\kappa(e)$ defined on edges; and sectional curvature $\rho(v_1, v_2, v_3)$ defined on vertex triples.
Some of these definitions can be extended to transition from one type to another (see Section~\ref{sec3.4}).
Finally, we review the concept of \textit{Ricci flow}, a geometric process that governs the evolution of spaces based on Ricci curvature. Several discrete analogues of Ricci flow have been proposed, and have found important applications in geometric data analysis and graph-based machine learning.

\begin{table}[!ht]
\caption{\noindent \textbf{Summary of discrete curvatures covered in this review.}
We provide detailed definitions of seven discrete curvature notions in this review.
This table summarizes the mathematical notations used to express them, their core theoretical motivations, and key references for further reading.
The seven notions of discrete curvature are categorized into three types: scalar curvature $p(v)$ defined on vertices; Ricci curvature $\kappa(e)$ defined on edges; and sectional curvature $\rho(v_1, v_2, v_3)$ defined on vertex triples.
}
\label{Tab:AllCurvs}
\scriptsize
\renewcommand{\arraystretch}{1.6}
\hyphenpenalty=10000\exhyphenpenalty=10000
\begin{tabular*}{\textwidth}{p{1.7cm}p{1.5cm}p{6.3cm}p{1.8cm}}
\toprule
\textbf{Curvature} & \textbf{Notation} & \textbf{Theoretical Motivation} & \textbf{References} \\
\midrule
Forman & $\kappa_F(e)$ & Based on the relationship between curvature and Riemannian Laplace operator & \cite{forman_bochners_2003, sreejith_forman_2016, sreejith_systematic_2017, weber2017characterizing, samal_comparative_2018} \\
Ollivier & $\kappa_O(e)$ & Based on the transportation distance between probability measures in metric spaces & \cite{ollivier_ricci_2007, ollivier_ricci_2009, ollivier2010survey, lin_ricci_2011, bauer_ollivierricci_2012, jost_olliviers_2014} \\
Bakry-\'Emery & $p_{BE}(v)$ & Provides a generalized lower bound of Ricci curvature obtained via the curvature-dimension inequality & \cite{bakry2006diffusions, lin_ricci_2010, jost_olliviers_2014, liu2018bakry, cushing2020bakry, mondal2024bakry} \\
Sectional & $\rho(v_1, v_2, v_3)$ & Quantifies the intersection patterns of closed balls in metric spaces & \cite{joharinad_topology_2019, joharinad_geometry_2022} \\
Menger & $\kappa_M(e)$ & Measures the radius of circumscribed circle of a triangle & \cite{menger1930untersuchungen, saucan2020simple, saucan2021simple} \\
Haantjes & $\kappa_H(e)$ & Measures the ratio between the length of an arc and the chord it subtends & \cite{haantjes1947distance, saucan2020simple, saucan2021simple} \\
Resistance & $p_{R}(v)$ & Quantifies the connectivity of a vertex in random spanning trees & \cite{devriendt2022effective, devriendt2022discrete, devriendt2024graph} \\
\bottomrule
\end{tabular*}
\end{table}

\subsubsection{Forman-Ricci Curvature}
\paragraph{\textbf{Computation on Graphs}}

Given a weighted graph \(\mathcal{G} = (\mathcal{V}, \mathcal{E}) \) where weights are assigned to both vertices and edges, the Forman-Ricci curvature of an edge $e \in \mathcal{E}$ connected by two vertices $v_1, v_2 \in \mathcal{V}$ is defined as follows:
$$
\kappa_F(e) =
w_e
\left(
    \frac{w_{v_1}}{w_e} + \frac{w_{v_2}}{w_e} - \sum_{e_{v_1} \sim e, \, e_{v_2} \sim e}
    \left[
        \frac{w_{v_1}}{\sqrt{w_e w_{e_{v_1}}}} + \frac{w_{v_2}}{\sqrt{w_e w_{e_{v_2}}}}
    \right]
\right),
$$
where \( e_{v_1} \sim e \) and \( e_{v_2} \sim e \) represent edges incident to \( v_1 \) and \( v_2 \) after excluding \( e \).

If the graph $\mathcal{G}$ is unweighted, we have $w_v = w_e = 1$ for all $v \in \mathcal{V}$ and $e \in \mathcal{E}$. In this case, the Forman-Ricci curvature reduces to the following simple and intuitive expression:
$$
\kappa_F^\#(e) = 4 - \text{deg}(v_1) - \text{deg}(v_2).
$$

Forman-Ricci curvature has been extensively applied in the analysis of several model and real-world networks including social networks, financial networks and biological networks  \cite{sreejith_forman_2016, samal_comparative_2018, samal_network_2021, chatterjee2021detecting}.
Several extensions of Forman-Ricci curvature have been proposed on graphs, including Forman-Ricci curvature of directed graphs \cite{saucan_discrete_2019} , as well as its augmentations that also consider cycles of arbitrary length \cite{fesser2024augmentations, ivanez2022comparative}.

\paragraph{\textbf{Computation on Simplicial Complexes}}

Given a weighted simplicial complex $\mathcal{K}$, the combinatorial Bochner-Weitzenböck formula introduced by Forman provides a method to assign a curvature function, denoted as \( \mathcal{F}_p(\sigma) \), to any \( p \)-simplex \( \sigma \) in a simplicial complex.
For edges (\( p = 1 \)), the resulting curvature function \( \mathcal{F}_1(\sigma) \) serves as a discrete analogue of the classical Ricci curvature.

Two simplices $\sigma_1, \sigma_2 \in \mathcal{K}_p$ are said to be \textit{parallel} (denoted as $\sigma_1 \parallel \sigma_2$) if and only if exactly one of the following condition holds:
(1) there exists a simplex $\tau \in \mathcal{K}_{p+1}$ such that $\sigma_1, \sigma_2 \subset \tau$, or
(2) there exists a simplex $\eta \in \mathcal{K}_{p-1}$ such that $\eta \subset \sigma_1, \sigma_2$.
In other words, two simplices of the same dimension are parallel if they share a coface (higher dimensional simplex) or a face (lower dimensional simplex), but not both.
We denote the set of parallel neighbors of a simplex $\sigma$ as $\mathrm{Parallel}(\sigma)$.

\begin{figure*}[h]
\begin{center}
\includegraphics[width = 0.8\textwidth]{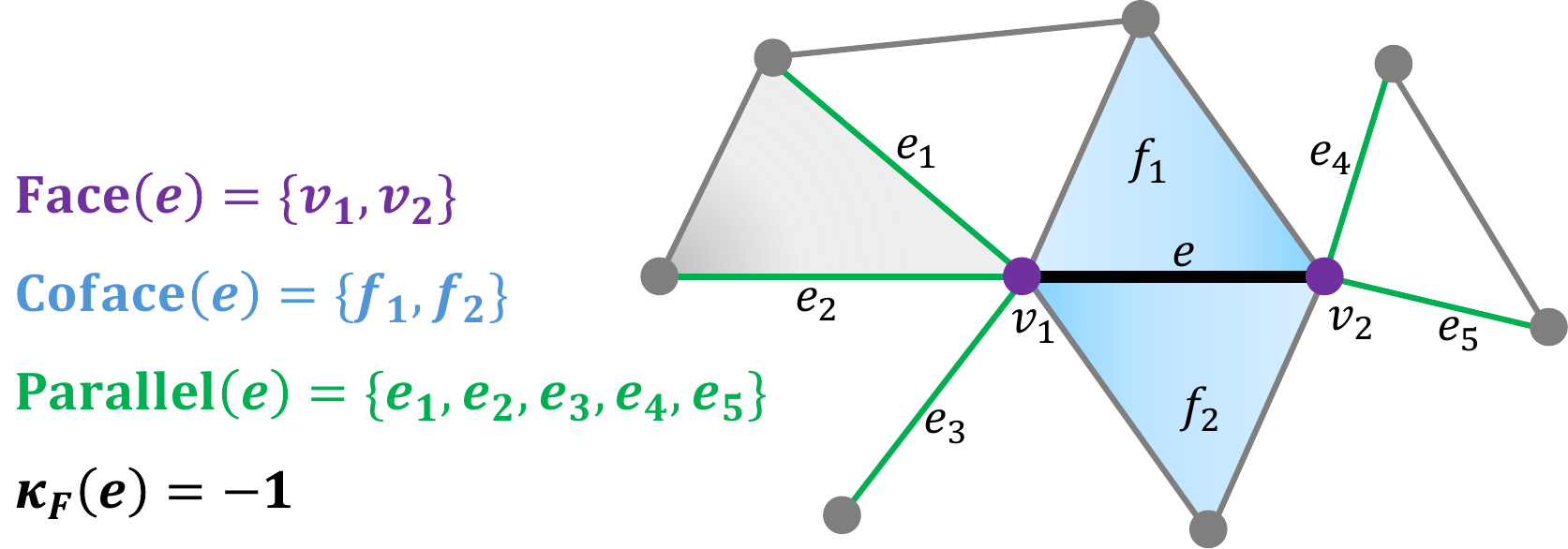}
\end{center}
\caption{
\textbf{Computation of Forman-Ricci curvature for an edge $e$ in an unweighted simplicial complex.}
The computation involves identifying the faces of $e$, which are the vertices at the ends of $e$; the cofaces of
$e$, which are the 2-simplices or triangles containing $e$; and parallel edges, defined as edges that share either a common face or a common coface with $e$, but not both.}
\label{Fig:FRC1}
\end{figure*}

\noindent The \textit{Forman-Ricci curvature} of an edge (or a 1-simplex) $e \in \mathcal{K}_1$ is defined as follows:
\begin{multline*}
\kappa_F(e) =
w_e
\Bigg[
    \left(
        \sum_{f \in \mathrm{Coface(e)}} \frac{w_e}{w_f}
        + \sum_{v \in \mathrm{Face}(e)} \frac{w_v}{w_e}
    \right) \\
    - \sum_{\hat{e} \in \mathrm{Parallel}(e)}
    \left|
        \sum_{f \in \mathcal{K}_2 : \hat{e}, e \subset f} \frac{\sqrt{w_e \cdot w_{\hat{e}}}}{w_f}
        - \sum_{v \in \mathcal{V} : v \subset \hat{e}, e} \frac{w_v}{\sqrt{w_e \cdot w_{\hat{e}}}}
    \right|
\Bigg].
\end{multline*}
If  $\mathcal{K}$ is unweighted, we have $w_{\sigma} = 1 \ \forall \sigma \in \mathcal{K}$.
In this case, the Forman-Ricci curvature reduces to the following simple combinatorial expression:
$$
\kappa^{\#}_F(e) = \#\mathrm{Face(e)} + \#\mathrm{Coface}(e) - \# \mathrm{Parallel}(e)
$$
The above definition of Forman-Ricci curvature has been applied to different types of abstract simplicial complexes, including clique complexes constructed from network data \cite{samal_comparative_2018, chatterjee2021detecting} and Vietoris-Rips complex constructed from point cloud data \cite{wee2021forman}.  Figure \ref{Fig:FRC1} shows an example computation of the Forman-Ricci curvature for an edge in a simplicial complex.

\paragraph{\textbf{Computation on Cubical Complexes and Hypergraphs}}

As noted in Section~\ref{sec2}, Forman's discretization of Ricci curvature is defined on weighted cell complexes.
Notably, cell complexes generalize both simplicial complexes and cubical complexes, differing primarily in the type of building blocks (or cells) they employ.
Simplices are the building blocks of simplicial complexes whereas elementary cubes are the building blocks of cubical complexes.
Therefore, the Forman-Ricci curvature of an edge in a cubical complex can be defined using the same notion of parallelism defined for simplicial complexes, with the distinction that the 2-dimensional faces in the case of cubical complexes are squares (or elementary cubes in $\mathbb{R}^2$).
Interested readers are referred to \cite{saucan2009combinatorial} for a detailed discussion on Forman-Ricci curvature for cubical complexes and its applications in image processing.

In the case of hypergraphs, several generalizations of Forman-Ricci curvature exist in literature, most of which are defined on directed hypergraphs \cite{leal2019curvature, leal2020ricci, saucan2019forman, leal2021forman}.
For the case of undirected hypergraphs, Forman-Ricci curvature has been defined via posets \cite{yadav2022poset} as well as by using a notion of Forman-Ricci curvature based on the Hodge Laplacian \cite{murgas2022hypergraph}.

\subsubsection{Ollivier-Ricci Curvature}
\paragraph{\textbf{Computation on Graphs}}

Let $\mathcal{G} = (\mathcal{V}, \mathcal{E})$ be a graph with vertex set $\mathcal{V}$ and edge set $\mathcal{E}$.  As discussed in Section~\ref{sec2}, Ollivier--Ricci curvature quantifies the cost of transporting a probability measure from one point in a metric space to another, capturing how the local geometry deviates from that of a flat space. Therefore, to compute the Ollivier--Ricci curvature of an edge in $\mathcal{G}$, it is necessary to define a probability measure associated with each vertex.
Given a vertex $v_i \in \mathcal{V}$ of degree $k$, and a set of its neighbors $\mathcal{N}_{v_i}$ the probability measure $m_i^{\alpha}$ associated with $v_i$, parameterized by $\alpha \in [0,1]$, is defined as follows:
$$
m_i^{\alpha}(v) =
\begin{cases}
\alpha & \text{if } v = v_i, \\
\frac{1 - \alpha}{k} & \text{if } v \in \mathcal{N}_{v_i}, \\
0 & \text{otherwise}.
\end{cases}
$$
In other words, we assign a discrete probability distribution centered at $v_i$, where a mass $\alpha$ is retained at the vertex itself and the remaining $(1 - \alpha)$ is equally distributed among its neighbors.

Let $d(v_i, v_j)$ denote the distance between vertices $v_i, v_j \in \mathcal{V}$.  If the graph $\mathcal{G}$ is unweighted, then $d(v_i, v_j)$ corresponds to the number of edges along the shortest path connecting $v_i$ and $v_j$. The Ollivier--Ricci curvature of an edge $e \in \mathcal{E}$ connecting vertices $v_i$ and $v_j$ is defined by the optimal transport problem:
$$
\kappa_O(e) = 1 - \frac{W_1(m_i^{\alpha}, m_j^{\alpha})}{d(v_i, v_j)},
$$
where $W_1$ denotes the Wasserstein distance between $m_i^{\alpha}$ and $m_j^{\alpha}$, given by
$$
W_1(m_i^\alpha, m_j^\alpha) = \inf_{\mu_{ij} \in \prod(m_i^\alpha, m_j^\alpha)} \sum_{(v_k, v_l) \in V \times V} d(v_k, v_l) \mu_{ij}(v_k, v_l).
$$
Here, $\prod(m_i^\alpha, m_j^\alpha)$ represents the set of joint probability measures $\mu_{ij}$ that satisfy the following conditions.
$$
\sum_{v_l \in V} \mu_{ij}(v_k, v_l) = m_i^\alpha(v_k); \quad \sum_{v_k \in V} \mu_{ij}(v_k, v_l) = m_j^\alpha(v_l).
$$

\begin{figure*}[h]
\begin{center}
\includegraphics[width = 0.8\textwidth]{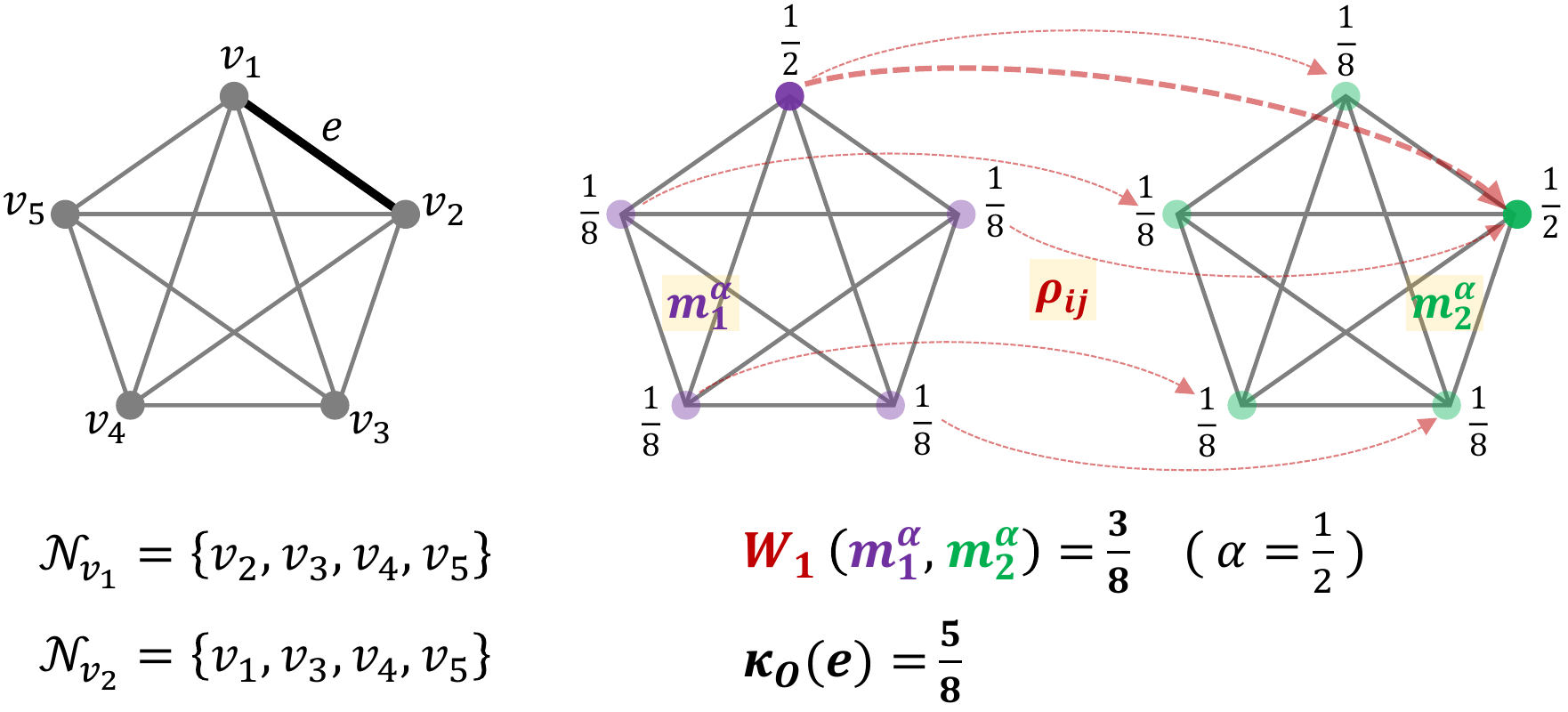}
\end{center}
\caption{
\textbf{Computation of Ollivier--Ricci curvature for an edge in a complete graph with five vertices.}
The edge $e = (v_1, v_2)$ is selected for curvature computation. The corresponding neighborhoods are identified as $\mathcal{N}_{v_1} = \{v_2, v_3, v_4, v_5\}$ and $\mathcal{N}_{v_2} = \{v_1, v_3, v_4, v_5\}$.
Discrete probability measures $m_1^\alpha$ and $m_2^\alpha$ are constructed using $\alpha = \frac{1}{2}$, such that each vertex retains half of the mass and distributes the remaining half uniformly across its neighbors (each neighbor receives $\frac{1}{8}$).
The right side of the figure displays the optimal transport plan $\rho_{ij}$ from $m_1^\alpha$ to $m_2^\alpha$.
The red dashed arrows represent the flow of probability mass between the two measures. The thickness of each arrow is proportional to the amount of mass transferred.
}
\label{Fig:ORC}
\end{figure*}

The above formulation considers all possible ways to transport the probability distribution $m_i^\alpha$ to $m_j^\alpha$ and identifies the one with the least cost, which is captured by the Wasserstein distance $W_1(m_i, m_j)$.
As a result, Ollivier-Ricci curvature compares the minimal transportation cost of distributing a mass over the neighbors of $v_i$ and $v_j$ with the distance between $v_i$ and $v_j$ itself.
Figure~\ref{Fig:ORC} illustrates this process on an edge in a complete graph with five vertices.
The probability measures are assigned based on the local neighborhoods of two vertices connected by the edge, and the optimal transport plan is visualized to show how mass is redistributed from one measure to the other.
Like Forman-Ricci curvature, Ollivier-Ricci curvature can also be extended to directed graphs (see \cite{saucan_discrete_2019} for a detailed discussion).
\paragraph{\textbf{Computation on Hypergraphs}}

Several generalizations of Ollivier-Ricci curvature have been developed for undirected hypergraphs \cite{asoodeh2018curvature, coupette2022ollivier, hacquard2024hypergraph}
as well as directed hypergraphs \cite{leal2019curvature, leal2020ricci, eidi2020ollivier}.
For undirected hypergraphs, , Asoodeh et al. \cite{asoodeh2018curvature} proposed a definition based on a multi-marginal optimal transport problem applied to a naturally induced random walk. 
More recently, Coupette et al. \cite{coupette2022ollivier} introduced the ORCHID framework, a unifying approach for generalizing Ollivier-Ricci curvature to hypergraphs. This framework extends the original formulation by generalizing both the probability measures assigned to nodes and the Wasserstein distance (also referred to as the aggregation function). Many existing definitions of Ollivier-Ricci curvature on hypergraphs can be viewed as specific instances of the ORCHID framework, each corresponding to different choices of probability measures and aggregation functions.

\subsubsection{Bakry-\'Emery Curvature}
\paragraph{\textbf{Computation on Graphs}}

As noted in Section~\ref{sec2}, Bakry and \'Emery's lower bound of Ricci curvature can be computed on any metric space via the curvature dimension inequality, provided the a well-defined notion of Laplacian exists.
Therefore, in order to define Bakry-\'Emery curvature on graphs, it is necessary to define a notion of graph Laplacian.
Let $\mathcal{G}=  (\mathcal{V}, \mathcal{E})$ be an unweighted, undirected graph, and let $f: \mathcal{V} \to \mathbb{R}$ be a function defined on the vertices in $\mathcal{G}$.
The non-normalized Laplacian $\Delta$ at a vertex $v \in \mathcal{V}$ can be defined as:
$$
\Delta f(v) = \sum_{u \in \mathcal{N}_v} (f(u) - f(v)),
$$
where $\mathcal{N}_v$ denotes the set of neighbors of $v$.
The above notion of Laplacian allows for the computation of operators $\Gamma(f)$ and $\Gamma_2(f)$ (see Section~\ref{sec2}), leading to the following curvature dimension inequality at $n \to \infty$,
$$
\Gamma_2(f)(v) \geq K_v  \cdot\Gamma(f)(v).
$$
Here $K_v$ is the lower Ricci curvature bound on $v$. The Bakry-\'Emery curvature $p_{BE}(v)$ of the vertex $v$ is the largest value of $K_v$ that holds for all functions $f$ at $v$. Since Bakry-\'Emery curvature assigns a function on the vertices of a graph, it is a type of scalar curvature.

The computation of Bakry-\'Emery curvature can be formulated as a semidefinite programming problem, where the value of curvature at a vertex depends on the spectral properties of the $\Gamma_2$ matrix.
Notably, it can be shown \cite{cushing2020bakry} that $p_{BE}(v)$ depends solely on the local topology of the graph around vertex $v \in \mathcal{V}$.
Specifically, one starts by defining the 2-ball centered at $v$, denoted as $B_2(v) = \{ u \in \mathcal{V} : d(u, v) \leq 2 \}$.
The vertices in $B_2(v)$ can be partitioned into the 1-neighborhood $S_1(v) = \{ u \in \mathcal{V} : d(u, v) = 1 \}$ and the 2-neighborhood $S_2(v) = \{ u \in \mathcal{V} : d(u, v) = 2 \}$.
The \textit{punctured 2-ball}, denoted as $\dot{B}_2(v)$, is a subgraph containing vertices from $S_1(v) \cup S_2(v)$ (or $B_2(v)\backslash \{v\}$).
However, $\dot{B}_2(v)$ is not simply the induced subgraph on $S_1(v) \cup S_2(v)$.
Instead, it is defined to include all edges between nodes in $S_1(v)$ and all edges between $S_1(v)$ and $S_2(v)$. It also excludes all edges between nodes in $S_2(v)$ since these have no influence on $p_{BE}(v)$.
It is this precise local structure of $\dot{B}_2(v)$ that completely determines the value of the Bakry--\'Emery curvature at $v$.

\begin{figure*}[h]
\begin{center}
\includegraphics[width = 0.8\textwidth]{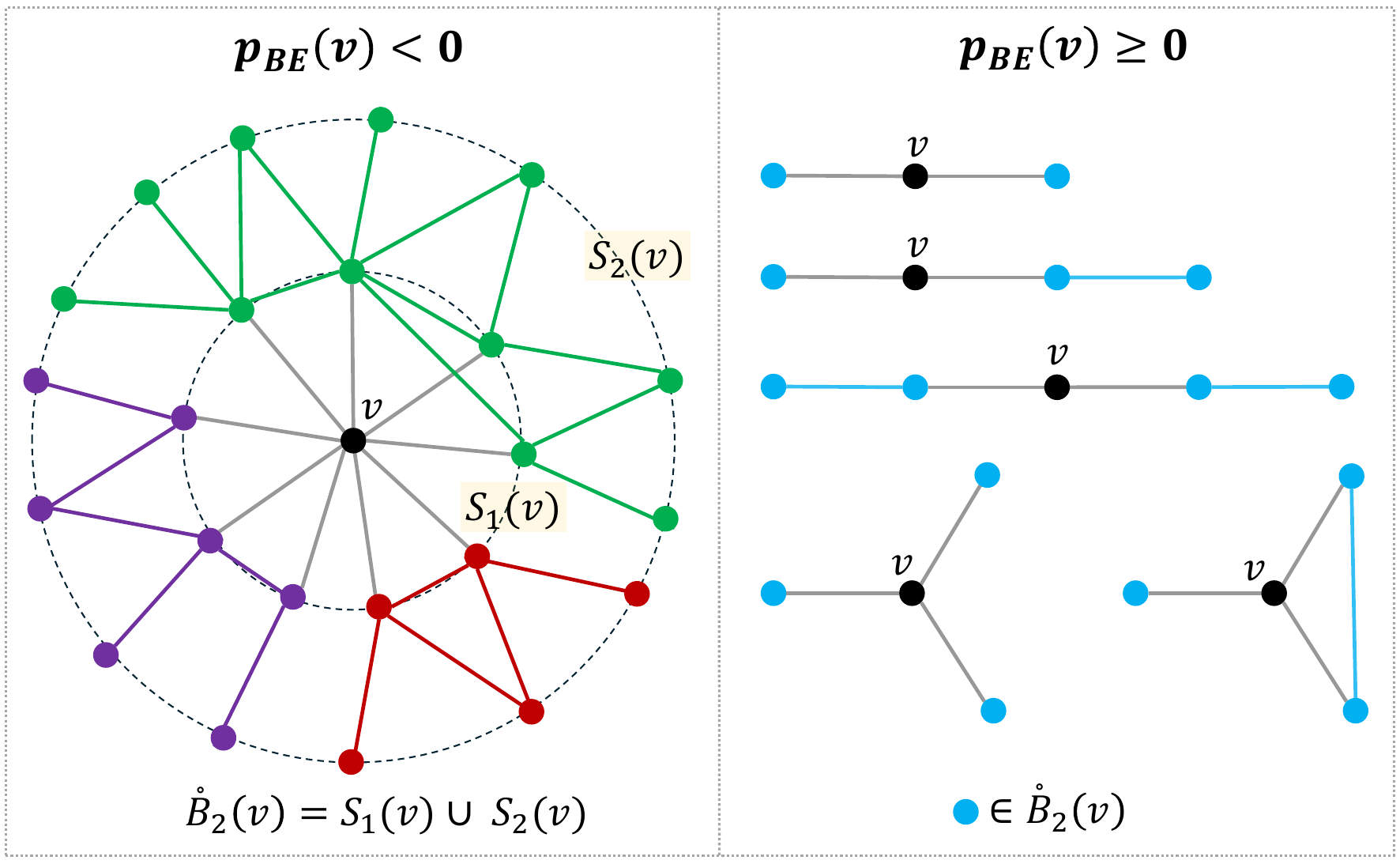}
\end{center}
\caption{
\textbf{Computation of Bakry--\'Emery curvature on graphs.}
The Bakry--\'Emery curvature $p_{BE}(v)$ at a vertex $v$ can be determined entirely based on the local structure of its punctured 2-ball $\dot{B}_2(v)$.
The punctured 2-ball consists of vertices in its 1-neighborhood $S_1(v)$ and 2-neighborhood $S_2(v)$, and includes all edges within $S_1(v)$ as well as those connecting $S_1(v)$ to $S_2(v)$, but excludes edges between nodes in $S_2(v)$.
On the left, Bakry-\'Emery curvature of the vertex $v$ is strictly negative due to the presence of multiple disconnected components in $\dot{B}_2(v)$, reflecting a structural bottleneck at $v$.
On the right, we show configurations that result in exceptions to this property, that is $p_{BE}(v) \geq 0$ even though $\dot{B}_2(v)$ is disconnected.
These cases are explicitly characterized in \cite{cushing2020bakry}.
}
\label{Fig:BERC}
\end{figure*}

The relationship between local topology and Bakry-\'Emery curvature is illustrated in Figure~\ref{Fig:BERC}, which shows examples of punctured 2-balls leading to either negative or non-negative curvature.
The curvature $p_{BE}(v)$ is strictly negative when the punctured 2-ball $\dot{B}_2(v)$ has more than two connected components, with only a few explicitly characterized exceptions.
These negatively curved vertices help characterize bottlenecks or sparsely connected regions in graphs, a phenomena which is also observed in Riemannian geometry.
For a formal investigation of the relationship between Bakry-\'Emery curvature and the local structure of graphs, we refer the reader to \cite{cushing2020bakry}. Empirical analyses of Bakry-\'Emery curvature on synthetic and real-world networks can be found in \cite{mondal2024bakry}.

\subsubsection{Sectional Curvature}
\paragraph{\textbf{Computation on graphs}}

The definition of sectional curvature provided in Section~\ref{sec2} can be easily extended to graphs, provided that a suitable notion of distance between vertices exists.
Given an unweighted, undirected, connected graph $\mathcal{G} = (\mathcal{V}, \mathcal{E})$, let $d(v_1, v_2)$ be the distance between between two vertices $v_1$ and $v_2$.
The \textit{sectional curvature} associated with any three vertices $v_1, v_2, v_3 \in \mathcal{V}$ is defined as
$$
\rho(v_1, v_2, v_3) := \inf_{v \in \mathcal{V}} \max_{i=1,2,3} \frac{d(v_i, v)}{r_i},
$$
where $r_i$'s are the Gromov products associated with closed balls $B_{r_i}(v_i) := \{ v \in \mathcal{V} : d(v_i, v) \leq r_i \}$ around vertices $v_1, v_2, v_3$.
For complete metric spaces, $1 \leq \rho(x_1, x_2, x_3) \leq 2$, where the lower bound is achieved for tripod spaces.
These bounds are also applicable to graphs.

\begin{figure*}[h]
\begin{center}
\includegraphics[width = 0.8\textwidth]{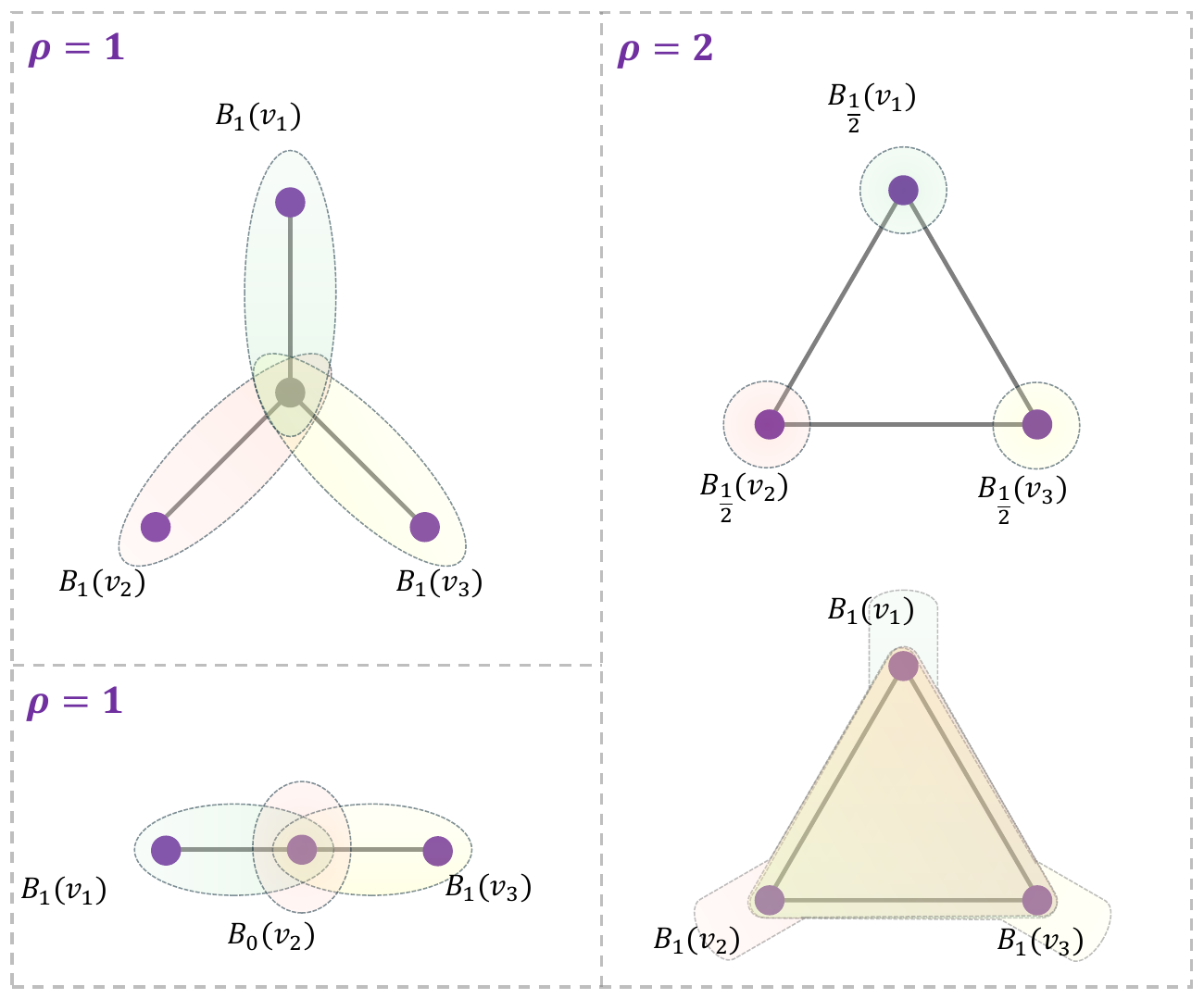}
\end{center}
\caption{
\textbf{Upper and lower bounds of sectional curvature on graphs.}
Sectional curvature \(\rho(v_1, v_2, v_3)\) is defined on a triple of vertices and quantifies the extent to which their neighborhood resembles a tree-like or fully connected structure.
The figure demonstrates the computation of sectional curvature $\rho$ on three graph configurations: a star graph, a path graph, and a complete graph.
In each case, closed balls $B_{r_i}(v_i)$ are initially assigned radii based on the Gromov products.
These balls are expanded simultaneously until a non-empty intersection is achieved.
The sectional curvature of the three vertices is the minimum value of the scaling factor at which this occurs.
In the star and path graphs (left), the closed balls intersect without expansion, yielding the lower bound $\rho = 1$.
In contrast, for the complete graph (right), the initial balls of radius $1/2$ do not intersect, and must be expanded to radius $1$, yielding the upper bound $\rho = 2$.}
\label{Fig:SC}
\end{figure*}

In Figure \ref{Fig:SC}, we demonstrate the computation of sectional curvature on three representative graph structures,
a star graph $S_4$, a path graph $P_3$, and a complete graph $K_3$. In both the star graph and the path graph, the lower bound of \(1\) is achieved.
The star graph $S_4$ consists of three leaves and one internal vertex. The curvature associated with the three leaves is equal to $1$. This is because the Gromov products yield closed balls of radius one centered at each leaf, and all three balls intersect at the central vertex. No further expansion is necessary to obtain a non-empty intersection.
In the path graph $P_3$ containing three vertices, the end vertices have unit-radius balls, and the central vertex has a zero-radius ball. These closed balls intersect at the central vertex, yielding curvature equal to $1$.
In contrast, the complete graph $K_3$ on three vertices yields the upper bound of sectional curvature.
Here, the Gromov products result in an initial radius of $\frac{1}{2}$ for each closed ball.
These balls do not intersect, hence they must be expanded to radius $1$ in order to achieve a non-empty intersection.
Since the balls are scaled by a factor of $2$, the sectional curvature is also equal to $2$.

More generally, it can be shown that \textit{the lower bound holds for any three vertices in a tree, while the upper bound holds for any three vertices in a complete graph}. As a result, sectional curvature provides a quantitative measure of where a graph lies on the spectrum between tree-like and fully connected structures.

\subsubsection{Menger-Ricci Curvature}

\begin{figure*}[h]
\begin{center}
\includegraphics[width = 0.9\textwidth]{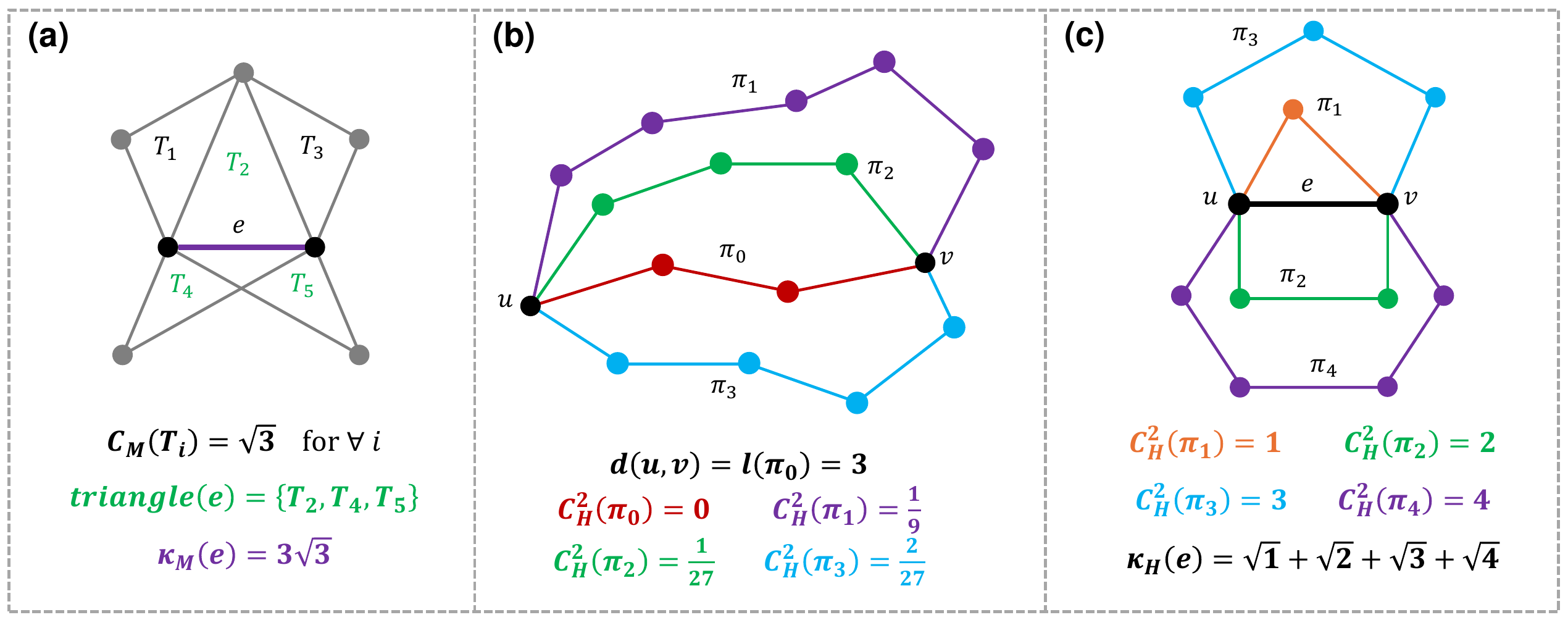}
\end{center}
\caption{
\textbf{Computation of Menger-Ricci and Haantjes-Ricci curvature on graphs}
\textbf{(a)} Example computation of Menger-Ricci curvature for an edge.
The curvature is computed by aggregating the Menger curvatures of all triangles adjacent to the edge.
\textbf{(b)} Example computation of Haantjes curvature for a path in the graph, treating the path as a discrete analogue of an arc and the shortest path between its endpoints as a chord.
\textbf{(c)} Computation of Haantjes-Ricci curvature for an edge.
In this case, the curvature is obtained by aggregating Haantjes curvatures over all the paths connecting the endpoints of the edge.}
\label{Fig:MRC-HRC}
\end{figure*}

Menger \cite{menger1930untersuchungen} proposed one of the simplest definitions of the curvature of a triangle in general metric spaces as the reciprocal of the radius of the circumscribed circle of the triangle.
In particular, given a metric space $(X,d)$ and a triangle $T$ with sides $a,b$ and $c$, the curvature of $T$ can be given by:
$$
C_M(T) = \frac{4 \sqrt{p(p-a)(p-b)(p-c)}}{a b c},
$$
where $p=(a+b+c)/2$ is the semiperimeter of $T$. The Menger curvature for a triangle is always positive.
Since Menger's definition of curvature is defined for general metric spaces, it can also be extended to graphs where the metric is induced by shortest path lengths.
\paragraph{\textbf{Computation on Graphs}}
Given an unweighted, undirected graph $\mathcal{G} = (\mathcal{V}, \mathcal{E})$, any triangle $T$ in $\mathcal{G}$ consists of three sides of unit length.
As a result, Menger curvature for any such triangle is given by $C_M(T)=\sqrt{3}$.
Since triangles in graphs can be interpreted as discrete analogues of two-dimensional sections or planes, the Menger curvature of any triangle $T$ can be considered as a type of sectional curvature.
However, unlike the definition of sectional curvature discussed earlier, which is based on the intersection patters on expanding closed balls, Menger's definition is highly localized.
Further, since all triangles in an unweighted graph have the same Menger curvature, this measure is not informative on its own.
A more meaningful approach \cite{saucan2020simple, saucan2021simple} is to consider the \textit{Menger-Ricci curvature} of edges, which is based on aggregating the curvatures of all triangles adjacent to an edge.
This aggregation is inspired by the differential geometric framework, where Ricci curvature is obtained by averaging sectional curvatures over planes along a given vector.
Formally, the Menger-Ricci curvature of an edge $e \in \mathcal{E}$ is defined as
$$
\kappa_M(e) = \sum_{T \in T_e} C_M(T) = \sqrt{3} \cdot |T_e|,
$$
where $T_e$ denotes the set of triangles in $G$ containing $e$.
An edge has a high positive Menger-Ricci curvature if it belongs to several triangles.
Figure \ref{Fig:MRC-HRC}(a) shows an example computation of Menger-Ricci curvature for an edge in a graph.

\subsubsection{Haantjes-Ricci Curvature}

Haantjes \cite{haantjes1947distance} proposed to measure the curvature of an arc in general metric spaces by comparing its length to that of the chord it subtends. Formally, given a curve in a metric space $(X,d)$, and three close enough points $p, q, r$ on this curve such that $p$ lies between $q$ and $r$, the curvature at point $p$ can be defined using the following expression:
$$
C_H^2(p) = 24 \lim_{q, r \to p} \frac{l(\overset{\frown}{qr}) - d(q, r)}{(d(q, r))^3},
$$
Here, $l(\overset{\frown}{qr})$ denotes the length of the arc $\overset{\frown}{qr}$ in the intrinsic metric induced by $d$. The Haantjes curvature of a path is always positive.
Building on this formulation, Saucan et al. \cite{saucan2020simple, saucan2021simple} developed a discrete analogue for the curvature of paths in graphs, and subsequently introduced a notion of \textit{Haantjes-Ricci curvature} of an edge.
\paragraph{\textbf{Computation on Graphs}}
Given an unweighted, undirected graph $\mathcal{G} = (\mathcal{V}, \mathcal{E})$, let $\pi = (v_0, v_1, \dots, v_n)$ be a path connecting vertices $v_0, v_n \in \mathcal{V}$.
Following the approach proposed by Haantjes, the path $\pi$ is treated as a discrete analogue of an arc, while the shortest path between $v_0$ and $v_n$ represents the corresponding chord.
Ignoring the proportionality constant, the Haantjes curvature of the path $\pi$ is then defined as:
$$
C_H^2(\pi) = \frac{l(\pi) - d(v_0, v_n)}{d(v_0, v_n)^3}
$$
Here, $l(\pi)$ denotes the length of the path, and  $d(v_0, v_n)$ represents the shortest path distance between $v_0$ and $v_n$.
If $v_0$ and $v_n$ form an edge $e \in \mathcal{E}$, then $d(v_0,v_n)=1$, and the resulting curvature, $C_H = \sqrt{n-1}$.
Consequently, the \textit{Haantjes-Ricci curvature} of the edge $e$ can is defined as
$$
\kappa_H(e) = \sum_{\pi \in \pi_e} C_H(\pi),
$$
where $\pi_e$ denotes all paths connecting the vertices $v_0$ and $v_n$ anchoring the edge $e$.
Note that while Menger-Ricci curvature accounts only for triangles (or paths of length $2$ between the vertices anchoring an edge), Haantjes-Ricci curvature also considers longer paths.
Figure \ref{Fig:MRC-HRC}(b) demonstrates the computation of Haantjes curvature of a path in a graph, whereas Figure \ref{Fig:MRC-HRC}(c) demonstrates the computation of Haantjes-Ricci curvature of an edge in a graph.

\subsubsection{Resistance Curvature}

\begin{figure*}[h]
\begin{center}
\includegraphics[width = 0.7\textwidth]{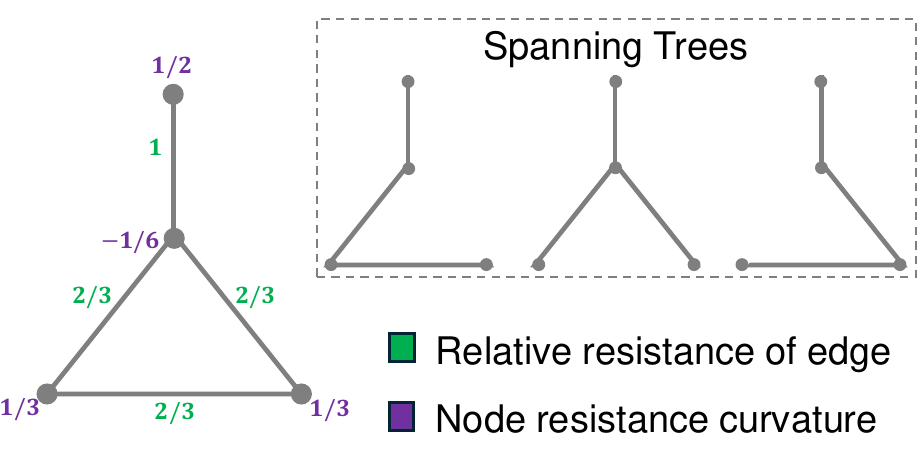}
\end{center}
\caption{\textbf{Relationship between random spanning trees, relative resistance of edges, and resistance curvature of vertices.}
The example graph consists of a triangle connected to an isolated vertex via a single edge. All possible spanning trees of the graph are enumerated.
The relative resistance of an edge is equal to the likelihood that it appears in a random spanning tree.
The edge connecting the isolated vertex to the triangle has a high relative resistance since its removal breaks the connectivity of the graph.
Edges within the triangle have lower relative resistance since their removal does not affect the connectivity of the graph.
The resistance curvature of a vertex is computed by aggregating the relative resistance across its incident edges.}
\label{Fig:RC}
\end{figure*}

All the reviewed discrete curvatures so far are rooted in Riemannian and metric geometry.
In contrast, Devriendt and Lambiotte \cite{devriendt2022discrete} introduced an alternative definition of curvature based on the notion of \textit{effective resistance}.
In the theory of electric circuits, effective resistance captures the influence of an entire network of resistors on the flow of current  between any two vertices \cite{thomassen1990resistances, dorfler2018electrical}.
As a result, effective resistance is closely linked with the connectivity of the underlying network.
Moreover, an important property of effective resistance is that it defines a valid metric on the set of nodes in a graph \cite{gvishiani1987metric}.
The relationship between effective resistance and graph topology, and its interpretation as a metric naturally leads to a definition of \textit{resistance curvature} on graphs.
\paragraph{\textbf{Computation on graphs}} 
Let $\mathcal{G} = (\mathcal{V}, \mathcal{E})$ be a weighted, undirected, connected graph where all the edge weights are positive.
If the vertex set is indexed as $\mathcal{V} = \{ v_1, v_2, ..., v_N\}$, then $w_{ij} \in \mathbb{R}^+$ denotes the weight of the edge between vertices $v_i, v_j \in \mathcal{V}$.
The Laplacian matrix $\mathbf{L}$ of the graph is defined as follows,
$$
\mathbf{L}_{ij} =
\begin{cases}
    -w_{ij} & \text{if } v_j \in \mathcal{N}_{v_i}, \\
    \text{deg}(v_i)     & \text{if } i = j, \\
    0       & \text{otherwise},
\end{cases}
$$
The \textit{Moore--Penrose pseudoinverse} $\mathbf{L}^\dagger$ of the Laplacian is determined by the equations $\mathbf{L}^\dagger \mathbf{L} = \mathbf{L} \mathbf{L}^\dagger = \mathrm{proj}(\ker(\mathbf{L})^\perp)$. It can be calculated by inverting the nonzero eigenvalues of the Laplacian matrix.
The \textit{effective resistance} between any two vertices $v_i, v_j \in \mathcal{V}$ is defined as
$$
\Omega_{ij} = (\mathbf{e}_i - \mathbf{e}_j)^\top \mathbf{L}^\dagger (\mathbf{e}_i - \mathbf{e}_j),
$$
where $\mathbf{e}_i$ is the $i$th unit vector.
Consequently, the resistance curvature of the vertex $v_i$ is defined as
$$
p_R(v_i) = 1 - \frac{1}{2} \sum_{v_j \in \mathcal{N}_{v_i}} \Omega_{ij} \cdot w_{ij}.
$$
Since resistance curvature assigns a function on the vertices of a graph, it is a type of scalar curvature.
The value of resistance curvature at a vertex depends on the quantity $\Omega_{ij} \cdot w_{ij}$, which is also known as the \textit{relative resistance} of the edge $(v_i,v_j)$.
This quantity reflects the structural importance of the edge within the connectivity of the graph.
Specifically, a redundant edge connecting two nodes within a densely interconnected region will have a low relative resistance.
In contrast, an important edge, whose removal would substantially impair the connectivity of the graph, will exhibit high relative resistance.
This property can be more formally understood by examining the relationship between relative resistance and \textit{random spanning trees} \cite{burton1993local, drineas2010effective, spielman2011spectral}.
A spanning tree on the graph $\mathcal{G}$ is a subset its edges that connects all vertices without forming cycles.
A random spanning tree $\mathcal{T}$ of $\mathcal{G}$ is selected from the set of all spanning trees of $\mathcal{G}$ with probability proportional to the product of the weights of its edges.
It can be shown that the relative resistance of an edge $(v_i, v_j)$ equals the probability that this edge is included in a random spanning tree, that is
$\Omega_{ij} \cdot w_{ij} = \Pr[(v_i, v_j) \in \mathcal{T}].$
Figure \ref{Fig:RC} demonstrates the relationship between random spanning trees, relative resistance, and resistance curvature.

\subsubsection{Extrinsic Curvature}

All the notions of discrete curvature reviewed so far are fundamentally \textit{intrinsic}; that is, they do not rely on any assumption about the embedding of the data in an ambient space.
Specifically, when defining curvature on graphs, the metric was derived entirely from graph connectivity, such as shortest path length, Wasserstein distance, or effective resistance, rather than from distances between vertices in an embedding space.
In contrast, several approaches to curvature in geometric data analysis adopt an extrinsic perspective \cite{aste2005complex, kamtue2018combinatorial, sritharan2021computing, ache2022approximating, jones2024manifold}.
These methods assume that the data is embedded in a higher-dimensional ambient space.
Any notion of curvature is then defined with respect to this embedding.
In this section, we briefly review two such notions of extrinsic curvature, namely combinatorial curvature \cite{kamtue2018combinatorial} and curvature derived from diffusion geometry \cite{jones2024manifold}.
\subsubsection*{\textit{Combinatorial Curvature}}

Combinatorial curvature is an extrinsic notion of discrete curvature that is defined on \textit{planar graphs}.
A graph $\mathcal{G} = (\mathcal{V}, \mathcal{E})$ is said to be planar if it can be embedded in $\mathbb{R}^2$ in such a way that its edges intersect only at the vertices.
In other words, it can be drawn on a plane in such a way that no two edges cross each other.
A planar graph partitions $\mathbb{R}^2$ into connected regions bordered by the edges that form a cycle.
Each connected region forms a polygon in $\mathbb{R}^2$ and is also known as a \textit{face}.
Combinatorial curvature is a type of scalar curvature, meaning it is defined on the vertices in $\mathcal{G}$.
Its value at a vertex $v \in \mathcal{V}$ depends on the degree of the vertex, $\text{deg}(v)$, and the degrees of faces $f \ni v$ incident to $v$, $\text{deg}(f)$. It can be expressed by the following formula:
$$
p_C(v) = 1 - \frac{\text{deg}(v)}{2} + \sum_{f \ni v} \frac{1}{\text{deg}(f)}.
$$

The above definition of combinatorial curvature reflects the \textit{angular defect}.
Since each face $f$ is a regular polygon of equal side length, the inner angle of $f$ would be $1 - \frac{2 \pi}{\text{deg}(f)}$.
Then, the sum of angles of all faces incident to the vertex $v$ would be:
$2\pi (1 - p_C(v)).$
If $p_C(v) < 0$, then the sum of angles at $v$ exceeds $2\pi$, indicating that the negihborhood of $v$ form a \textit{saddle-shaped} (hyperbolic) surface locally.
Conversely, if $p_C(v) > 0$, the sum of angles is less than $2\pi$, and the neighborhood of $v$ resembles an \textit{elliptical} (spherical) surface.
When $p_C(v) = 0$, the local geometry around $v$ is \textit{flat} (Euclidean).
Thus, the sign of combinatorial curvature reflects the local geometry around a vertex.

\subsubsection*{\textit{Diffusion Geometry-based Curvature}}

Jones \cite{jones2024manifold} provided a framework based on \textit{diffusion geometry} \cite{jones2024diffusion} to estimate the curvature of a manifold $\mathcal{M}$ embedded in Euclidean space $\mathbb{R}^D$, using a set of discrete points sampled from $\mathcal{M}$.
This approach relies on the carré du champ formula, which expresses $g(\nabla f, \nabla h)$ in terms of the Laplacian $\Delta$.
The carré du champ formula was introduced earlier in the discussion of Bakry-\'Emery's lower bound on Ricci curvature (Section~\ref{sec2}).
In the rest of the subsection, we use $\Gamma_c(f,h) := g(\nabla f, \nabla h)$ to denote the \textit{carré du champ operator}.

The carré du champ operator can be used to obtain the tangent space $\mathcal{T}_p\mathcal{M}$ at a point $p \in \mathcal{M}$, using a $D \times D$ gram matrix $\mathbf{G}(p)$ of the inner products of the gradients of the coordinate functions $\{\nabla_p x_i : i=1,...,D \}$,
$$
\mathbf{G}_{ij}(p) = g(\nabla_p x_i, \nabla_p x_j) = \Gamma_c(x_i, x_j).
$$
The eigenvectors of $\mathbf{G}(p)$ form an orthonormal basis for $\mathbb{R}^D$, such that the top $d$ eigenvectors form an orthonormal basis $\nabla_p x_1, \ldots, \nabla_p x_d$for the tangent space $T_p \mathcal{M}$, while the remaining $D - d$ form an orthonormal basis $\vec{n}^{1}, \ldots, \vec{n}^{D-d}$ for the normal space.
Finally, the curvature can be expressed in terms of the Hessian $\alpha^{\ell}_{ij} = H_p(n^\ell)(\nabla_p x_i, \nabla_p x_j)$ for each normal vector $n^\ell$ and each pair of tangent directions $\nabla_p x_i, \nabla_p x_j$,
$$
H(f)(\nabla h_1, \nabla h_2) = \frac{1}{2} \left( \Gamma(h_1, \Gamma(h_2, f)) + \Gamma(h_2, \Gamma(h_1, f)) - \Gamma(f, \Gamma(h_1, h_2)) \right).
$$
Specifically, the Riemannian curvature tensor, Ricci curvature and scalar curvature, respectively, are given by
$$
R_{ijkl} = \sum_{\ell=1}^{D-d} \left( \alpha_{ik}^\ell \alpha_{jl}^\ell - \alpha_{jk}^\ell \alpha_{il}^\ell \right),
$$
$$
\operatorname{Ric}_{ij} = \sum_{\ell=1}^{D-d} \sum_{k=1}^d \left( \alpha_{kk}^\ell \alpha_{ij}^\ell - \alpha_{ik}^\ell \alpha_{jk}^\ell \right),
$$
$$
S = \sum_{\ell=1}^{D-d} \sum_{i,j=1}^d \left( \alpha_{ii}^\ell \alpha_{jj}^\ell - (\alpha_{ij}^\ell)^2 \right).
$$

The overall strategy is to estimate the Laplacian $\Delta$ from data using the \textit{variable bandwidth diffusion kernel} method \cite{coifman2006diffusion, berry2016variable} and compute $\Gamma_c(f,h)$.
This estimate of $\Gamma_c$ can be used to infer the curvature of the manifold $\mathcal{M}$ via the Hessian.
Given a point cloud $X = \{p_1, \ldots, p_n\}$, the kernel matrix is defined as
$$
K_\epsilon(x_i, x_j) = \exp \left( -\frac{ \| x_i - x_j \|^2 }{ \epsilon \, \rho_b(x_i) \rho_b(x_j) } \right),
$$
where $\rho_b$ is an automatically defined bandwidth function, and $\epsilon$ is the bandwidth parameter.
The resulting kernel is used to construct an estimate of $\hat{\Delta}_\epsilon$ of the Laplacian operator, which in turn provides an estimate of the carré du champ operator,
$$
\hat{\Gamma}_c(f,h) = \frac{1}{2} \left[ f \hat{\Delta}_\epsilon h + h \hat{\Delta}_\epsilon f - \hat{\Delta}_\epsilon(fh) \right].
$$
This estimate of the carré du champ operator is used to compute $\hat{\mathbf{G}}(p_i)$ at each $p_i \in X$, along with the estimates of the Hessian terms $\hat{\alpha}^{\ell}_{ij}$.
Finally, the estimates of Riemannian curvature $\hat{R}_{ijkl}$, Ricci curvature $\hat{\operatorname{Ric}}_{ij}$, and scalar curvature $\hat{S}$ can be obtained.

\subsection{Discrete Ricci Flow for Data Analysis}\label{sec3.3}

In Riemannian manifolds, regions with large positive curvature tend to be more densely packed than regions of negative curvature.
To identify these regions of large curvature, Hamilton \cite{hamilton1982three} introduced Ricci flow, a diffusion process governed by curvature that deforms the manifold in order to smooth out the irregularities in the metric.
This process shares several similarities with the heat diffusion equation.
Formally, given a Riemannian manifold $M$ with the metric $g_{ij}$ and Ricci curvature $R_{ij}$, Hamilton's Ricci flow is expressed as
$$
\frac{\partial}{\partial t} g_{ij} = -2 R_{ij}. \nonumber
$$
The above equation suggests that regions with positive sectional curvature tend to shrink whereas regions of negative sectional curvature tend to expand and spread out. Ni et al. \cite{ni2019community} utilized Ollivier-Ricci curvature to introduce a discrete Ricci flow process, where the edge weights in a network are updated according to the following equation:
$$
w_{ij}^{(t+1)} = d^{(t)}(i, j) - \kappa_{ij}^{(t)} \cdot d^{(t)}(i, j). \nonumber
$$
Here $w_{ij}^{(t)}$ and $\kappa_{ij}^{(t)}$ are the weight and the Ollivier-Ricci curvature respectively of the edge between vertices $i$ and $j$ at the $t$-th time step, and $d^{(t)}(i, j)$ is the distance on the graph induced by the weights $w_{ij}^{(t)}$.
Further, $w_{ij}^{(0)} = w_{ij}$ and $d_{ij}^{(0)} = d_{ij}$. Similar to its Riemannian counterpart, the discrete Ricci flow process results in a increase in the weights of negatively curved edges and a decrease in the weights of positively curved edges.

\subsection{Multiscale Curvature for Data Analysis}\label{sec3.4}
\begin{table}[!ht]
\caption{\noindent \textbf{Ability of curvatures to characterize data at multiple scales.}
The table compares the applicability of different notions of discrete curvature, namely Forman (\textbf{F}), Ollivier (\textbf{O}), Bakry-\'Emery (\textbf{BE}), Sectional (\textbf{S}), Menger (\textbf{M}), Haantjes (\textbf{H}), and Resistance (\textbf{R}) to various geometric objects.
A big star ($\bigstar$) represents the original definition being directly applicable to the given geometric object, two smaller stars ($\star\star$) denote a well-defined extension in literature, a single small star ($\star$) indicates that an extension is possible but no established framework exists, and “--” signifies that the curvature cannot be applied or defined on the given object.
}
\scriptsize
\label{Tab:MultiscaleCurv}
\renewcommand{\arraystretch}{1.5}
\begin{tabular*}{\textwidth}{@{\extracolsep\fill}lccccccc}
\toprule
\textbf{Geometric Object} & \textbf{F} & \textbf{O} & \textbf{BE} & \textbf{S} & \textbf{M} & \textbf{H} & \textbf{R} \\
\midrule
Vertices                  & $\star\star$    & $\star\star$      & $\bigstar$           & $\star$       & $\star\star$     & $\star\star$       & $\bigstar$         \\
Edges                     & $\bigstar$      & $\bigstar$        & --                   & $\star$            & $\star\star$     & $\star\star$       & $\star\star$       \\
3-vertex cycles & --              & --                & --                   & $\bigstar$         & $\bigstar$       & --                 & --                 \\
Simplices                 & $\bigstar$      & --                & --                   & --                 & --               & --                 & --                 \\
Paths                     & --              & --                & --                   & --                 & --               & $\bigstar$         & --                 \\
3-vertex sets     & --              & --                & --                   & $\bigstar$         & $\star$          & --                 & --                 \\
$n$-vertex sets   & --              & --                & --                   & $\star$           & --               & --                 & --                 \\
\bottomrule
\end{tabular*}
\end{table}
\begin{figure*}[!ht]
\begin{center}
\includegraphics[width = \textwidth]{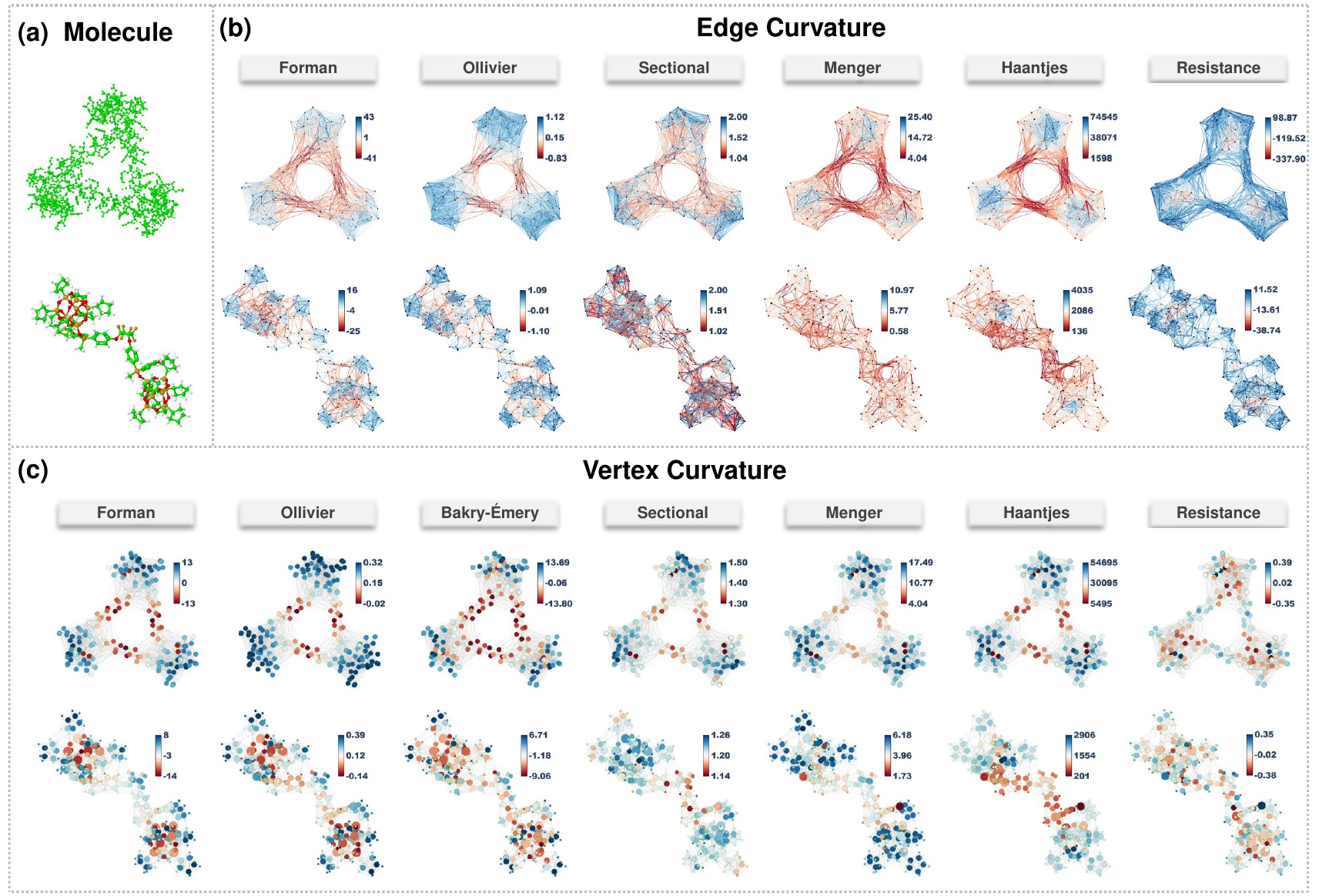}
\end{center}
\caption{\textbf{Comparison of different notions of discrete curvature on molecular systems.}
(a) Molecular structures used for illustration: the protein 3ULG (top) and the dimer representation of a polymer (bottom).
(b) Edge curvature visualizations for six different definitions: Forman-Ricci, Ollivier-Ricci, sectional, Menger-Ricci, Haantjes-Ricci, and resistance curvature.
(c) Node curvature visualizations for six definitions: Forman-Ricci, Ollivier-Ricci, Bakry-\'Emery, sectional, Menger-Ricci, Haantjes-Ricci, and resistance curvature.
For the protein, a geometric graph of C$_\alpha$ atoms was used with 15.0~\AA\ cutoff. For the dimer, all atoms were considered with 4.0~\AA\ cutoff.
Forman-Ricci curvature was computed on the clique complex of the geometric graph, while other curvature notions were directly computed on the geometric graphs.
Across both examples, curvature measures qualitatively differentiate between densely connected regions (clusters) and sparse connecting regions (bottlenecks).
In the dimer example, different measures provide complementary perspectives on the backbone, silicon-oxygen cage, and attached cyclopentane rings.}
\label{Fig:CurvCompare}
\end{figure*}

In the previous section, we reviewed seven definitions of discrete curvature, categorizing them into three broad types: scalar curvature defined on vertices, Ricci curvature defined on edges, and sectional curvature defined on vertex triples.
Notably, certain curvature models exhibit the flexibility to transition across categories.
For example, any Ricci-type curvature, although fundamentally defined on edges, can be used to characterize the curvature of vertices.
Moreover, some of notions of curvature can be generalized to higher-order geometric objects such as simplices, triangles or paths, offering rich avenues for exploring the geometry of complex datasets.
Finally, sectional curvature is a global geometric invariant as it is defined on any three vertices in the graph. However, it can be meaningfully defined at the level of vertices or edges, while preserving global information.
Table~\ref{Tab:MultiscaleCurv} summarizes the applicability of discrete curvatures to various geometric objects including vertices, edges, paths, simplices, and $n$-vertex combinations, highlighting the versatility of discrete curvature as a multiscale characterization tool.

Figure \ref{Fig:CurvCompare} illustrates the computation of different notions of curvature at the vertices (scalar curvature) and at the edges (Ricci curvature) for two molecular systems: the protein 3ULG and the dimer representation of a polymer.
For the protein, a geometric graph was constructed using the coordinates of C$_\alpha$ atoms, with a distance cutoff of 15.0~\AA.
For the dimer, the coordinates of all atoms were used, with a distance cutoff of 4.0~\AA.
All curvatures except Forman-Ricci curvature were computed directly on these geometric graphs.
Forman-Ricci curvature was computed on the clique complex derived from each graph.

Across both systems, the curvatures consistently distinguish between densely packed regions (clusters) and sparsely connected regions (bottlenecks).
Specifically, for Forman-Ricci curvature and Ollivier-Ricci curvature, vertices and edges within bottlenecks exhibited negative curvature, while those in clusters showed positive curvature.
A similar pattern was observed for the Bakry-\'Emery curvature computed on vertices.
Although Menger-Ricci, Haantjes-Ricci, and sectional curvature are non-negative by definition, they also effectively differentiated between clusters and bottlenecks.
Particularly, for sectional curvature, we discussed that trees correspond to the theoretical lower bound, and complete graphs correspond to the upper bound.
Bottlenecks resemble trees due to sparse connectivity and hence have lower curvature values, while clusters resemble cliques due to dense local connections, and hence have higher curvature values.

Application of different notions of discrete curvature to the protein revealed substantial similarities in their qualitative patterns.
However, each curvature highlights different geometric characteristics of the underlying data, and they should not be regarded as interchangeable.
This complementarity is particularly evident in the dimer example, which consists of a molecular backbone, a silicon-oxygen cage, and cyclopentane rings attached to the cage.
For atoms within the cage (silicon and oxygen), Forman-Ricci curvature, Ollivier-Ricci curvature, and Bakry-\'Emery Ricci curvature yield distinct curvature values for atoms in the cage.
These observations suggest that combining multiple curvature notions can provide complementary geometric insights from the underlying data.
A more detailed empirical analysis is required to systematically assess the strengths and limitations of each curvature measure.
Such an analysis is beyond the scope of this review.
However, readers are referred to \cite{samal_comparative_2018} and \cite{mondal2024bakry} for empirical comparisons of different notions of discrete curvature.

\subsubsection*{\textit{Forman-Ricci and Ollivier-Ricci curvatures are edge-based geometric invariants}}
In Riemannian geometry, Ricci curvature is a contraction of the curvature tensor along a tangent vector, yielding an invariant that characterizes volume growth in a given direction.
In discrete settings such as graphs or simplicial complexes, edges serve as analogues of tangent vectors in smooth manifolds \cite{samal_comparative_2018}.
Consequently, any discretization of curvature that assigns values to edges can be regarded as a form of \textit{discrete Ricci curvature}.
Forman-Ricci curvature is derived from a combinatorial analogue of the Bochner-Weitzenböck formula, while Ollivier-Ricci curvature is based on optimal transport of probability measures between the neighborhoods of adjacent vertices.
This edge-based formulation of Ricci-type curvatures is particularly relevant in the context of network data analysis \cite{eidi2020edge}.
Networks are inherently defined by the relationships (edges) between their components (vertices).
While many standard network measures such as degree centrality or clustering coefficient focus on quantifying vertex properties, discrete Ricci curvatures offer a complementary perspective by quantifying the connections themselves, thereby providing a relational perspective for understanding the geometry of networks.

\subsubsection*{\textit{Forman's discretization of curvature is highly flexible and generalizable}}
\begin{figure*}[!ht]
\begin{center}
\includegraphics[width = 0.6\textwidth]{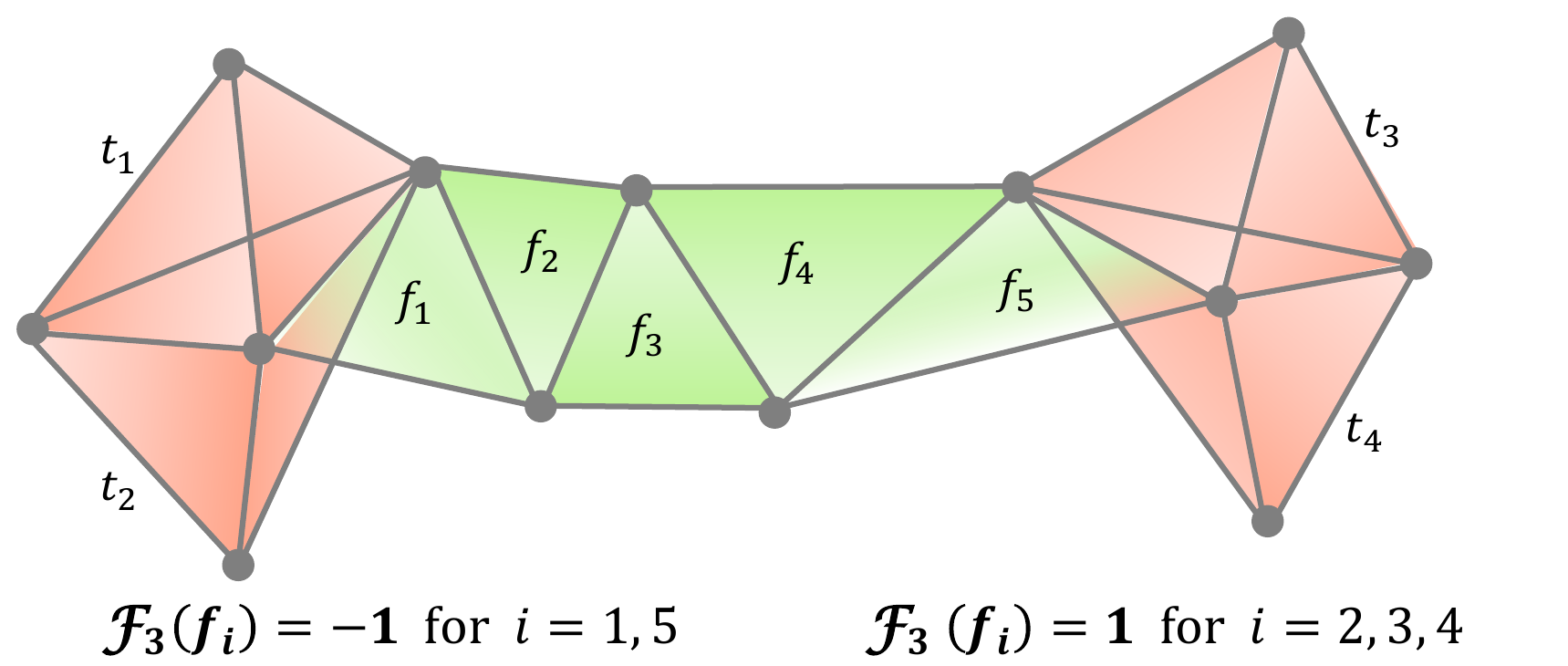}
\end{center}
\caption{\textbf{Computation of Forman curvature on 2-simplices in an unweighted simplicial complex.}
The figure shows a simplicial complex with five labeled 2-simplices $f_1, \ldots, f_5$ highlighted in green.
In addition, the simplicial complex includes four 3-simplices (tetrahedra), along with $14$ associated $2$-simplices that are not labeled in the figure.
The Forman curvature of each labeled $2$-simplex is indicated in the figure.
The simplices $f_1$ and $f_5$ have three faces, zero cofaces, and four parallel simplices, yielding a curvature value of $-1$.
The simplices $f_2, f_3$ and $f_4$ have three faces, zero cofaces and two parallel simplices, yielding a curvature value of $1$.}
\label{Fig:FRC2}
\end{figure*}

The Ricci-type curvature based on Forman's definition can characterize the geometry of edges in graphs as well as simplicial complexes (see Figure~\ref{Fig:FRC1}).
However, unlike other approaches to discrete curvature, Forman's approach provides a rigorous framework to define curvature for geometric objects of arbitrary dimension.
Specifically, given a weighted simplicial complex $\mathcal{K}$, the combinatorial Bochner--Weitzenböck formula defines a curvature function $\mathcal{F}_p(\sigma)$ for each $p$-simplex $\sigma \in \mathcal{K}$ ($\mathcal{F}_1(\sigma)$ corresponds to the curvature of an edge and serves as a discrete analogue of classical Ricci curvature).
This curvature function for a $p$-simplex depends on the weights of its faces, cofaces, and parallel simplices. These terms are formally defined in Section~\ref{sec3.1}.
In the case of unweighted simplicial complexes, the curvature function reduces to the following expression:
$$
\mathcal{F}_p^{\#}(\sigma) = \#\mathrm{Face}(\sigma) + \#\mathrm{Coface}(\sigma) - \#\mathrm{Parallel}(\sigma)
$$
This expression is directly applicable to any simplex $\sigma$ and generalizes the Ricci curvature formulation defined on edges.
Figure~\ref{Fig:FRC2} demonstrates the computation of Forman curvature for 2-simplices in a simplicial complex.
For a detailed treatment of Forman curvature in weighted simplicial complexes, we refer the reader to Forman's original work \cite{forman_bochners_2003}.

\subsubsection*{\textit{Menger and Haantjes curvatures characterize triangles and paths respectively}}
Menger and Haantjes curvatures originate in metric geometry where they are defined on higher-order geometric objects beyond individual points or pairs of points.
Menger curvature is defined for triangles in metric spaces and measures curvature as the reciprocal of the radius of the circumcised circle.
Haantjes curvature is defined on arcs and measures their deviation from straight lines by comparing arc length to the length of the chord it subtends.
These concepts can be naturally translated to the discrete setting of graphs.
In particular, Menger curvature can be defined on triangles or 3-cycles whereas Haantjes curvature can be defined on paths. They therefore provide a way to characterize the geometry of structures other than vertices or edges in a graph.
We previously discussed how both Menger and Haantjes curvatures can also be extended to edges.
These edge-based extensions, known as Menger-Ricci and Haantjes-Ricci curvature, allow the original definitions to be used within the framework of standard network analysis where quantifying the relationships between vertices is essential.

\subsubsection*{\textit{Ricci-type curvatures can characterize vertex-level geometry}}
In Riemannian geometry, scalar curvature is a further contraction of the curvature tensor and is roughly interpreted as the average of the Ricci curvatures across all tangent vectors at a given point.
It assigns a single real number to each point on a manifold, reflecting the geometry of the metric near that point.
Drawing on this analogy, one can define a \textit{discrete scalar curvature} for vertices in a network by averaging the Ricci curvatures of all edges incident to a given vertex.
For each of the four Ricci-type curvatures considered in this work, namely Forman-Ricci, Ollivier-Ricci, Menger-Ricci, and Haantjes-Ricci curvature, the scalar curvature of a vertex $v \in \mathcal{G}$ can be defined as
$$
p(v) = \frac{1}{\deg(v)} \sum_{e \in E(v)} \kappa(e)
$$
where $\deg(v)$ denotes the degree of vertex, and $E(v)$ is the set of edges incident to $v$, and $\kappa(e)$ represents the Ricci-type curvature assigned to edge $e$.
This vertex-level formulation provides a scalar summary of the local geometric environment around a vertex and serves as a direct analogue of scalar curvature in the Riemannian setting.

While the above averaging approach offers a straightforward analogue to the Riemannian setting, it remains largely heuristic and does not guarantee that $p(v)$ converges to the continuous scalar curvature in the manifold limit, as it treats all incident edges equally. 
A recent formulation \cite{hickok2025discrete} considers a weighted aggregation of Ollivier-Ricci curvature that can be shown to converge to the definition of scalar curvature on smooth manifolds. 
In particular, the \textit{scalar Ollivier-Ricci curvature} at a vertex $v \in \mathcal{G}$ can be defined as
$$
p_{\text{SORC}}(v) = \frac{1}{\deg(v)} \sum_{e \in E(v)} w(e)^2 \, \kappa_O(e)
$$
where $w(e)$ denotes the edge weight. 
According to this formulation, when vertices correspond to sampled points on a manifold and edge weights reflect their underlying geodesic distances, the weighting scheme naturally assigns higher importance to edges connecting more distant points.

\subsubsection*{\textit{Bakry--\'Emery and Resistance curvatures are fundamentally defined on vertices}}
Bakry--\'Emery curvature is fundamentally a quantity derived from Ricci curvature but is defined at the vertex level.
Its formulation is based on the curvature-dimension inequality, which establishes lower bounds for Ricci curvature.
In the Riemannian setting, this inequality is evaluated pointwise by considering all vectors in the tangent space of a given point.
In discrete spaces such as graphs, this naturally leads to a vertex-based definition where curvature is associated with a vertex rather than an edge.
In particular, the Bakry--\'Emery curvature at a vertex reflects the minimum Ricci curvature across all edges incident to that vertex.

Resistance curvature is also defined on vertices and is based on the concept of relative resistance.
The relative resistance of an edge is the probability that it appears in a randomly chosen spanning tree.
Edges with high relative resistance are more structurally important since their removal disrupts connectivity, whereas edges with low relative resistance are redundant in spanning trees.
The resistance curvature at a vertex is defined by aggregating the relative resistances of its incident edges.
It can be shown that it is related to the expected degree of the vertex in a random spanning tree. This interpretation of resistance curvature links it to the connectivity of random spanning trees.
In addition to the vertex-based formulation, which serves as a discrete analogue of scalar curvature, Devriendt and Lambiotte \cite{devriendt2022discrete} proposed an extension of resistance curvature to edges.
This edge-based formulation corresponds to a Ricci-type curvature, and is defined for an edge $e = (v_i, v_j) \in \mathcal{E}$ as follows:
$$
\kappa_R(e) = \frac{2 \cdot [ p_R(v_i) + p_R(v_j) ]}{\omega_{ij}}.
$$

\subsubsection*{\textit{Sectional curvature captures interactions among triples of vertices}}
Most definitions of discrete curvature are defined on local geometric objects such as vertices or edges. While some of these curvatures can be extended to higher-order structures, such as Forman curvature to higher-dimensional simplices, Menger curvature to triangles, and Haantjes curvature to paths, these extensions primarily capture local geometric information \cite{saucan2021simple}.
On the other hand, sectional curvature offers a global perspective by evaluating interactions among triples of vertices.
For any three vertices in a connected component of a graph, sectional curvature quantifies how much the geodesic balls centered at the three vertices must be enlarged for a common intersection to occur.
Intuitively, this definition captures the geometric structure enclosed by subgraph contained within these three points.

Although sectional curvature is fundamentally defined over vertex triples, it can also be extended to edges and vertices.
The sectional curvature at an edge can be defined as the average sectional curvature across all vertex triples that contain that edge,
$$
\kappa(e) = \frac{1}{|T_e|} \sum_{\{v_1,v_2,v_3\} \in T_e} \rho(v_1,v_2,v_3),
$$
where $T_e$ is the set of all vertex triples that include the edge $e \in \mathcal{E}$.
Similarly, vertex curvature can be defined by averaging sectional curvature across all triples that contain a given vertex,
$$
p(v) = \frac{1}{|T_v|} \sum_{\{v_1,v_2,v_3\} \in T_v} \rho(v_1,v_2,v_3),
$$
where $T_v$ is the set of all vertex triples containing $v\in \mathcal{V}$.
These definitions align with the discrete analogues of Ricci and scalar curvatures, respectively.
While such edge- and vertex-based formulations of sectional curvature have not been previously introduced in literature, they enable consistent comparison with other discrete curvatures.
Specifically, in our example computation on molecular graphs, we demonstrated that these quantities can help distinguish densely packed regions and bottlenecks and reveal global geometric patterns in molecules.
A detailed investigation of such applications is beyond the scope of this work and is left for future study.

\section{Curvature-based Geometric Learning}\label{sec4}

Among the various definitions of discrete curvature reviewed in this work, Forman-Ricci curvature and Ollivier-Ricci curvature have particularly demonstrated broad applicability in machine learning, spanning both supervised and unsupervised methods.
The majority of these applications involve graph-based machine learning.
Graphs have attracted significant attention within the machine learning community in recent years due to their capacity to model diverse relational systems \cite{sperduti1993encoding, goller1996learning, frasconi1998general, bruna2013spectral}.
Our review encompasses curvature-based geometric learning in three key areas, including community detection, manifold learning, and graph neural networks.

\subsection{Community Detection}
Community detection can be viewed as a type of unsupervised node clustering on graphs, which has several important applications in social network analysis, fraud detection, tumor identification and understanding the dynamics of disease spread \cite{karatacs2018application}.
A community is an intermediate or \textit{mesoscopic} structure consisting of a collection of vertices that are closely connected to one another but have sparse connections to other densely connected vertices in the network.
The problem of community detection has garnered significant interest of statistical physics, network science and machine learning communities, \cite{fortunato2010community, porter2009communities, abbe2018community}. Prominent approaches include methods based on spectral graph clustering \cite{krzakala2013spectral}, statistical inference methods \cite{boccaletti2007detecting, newman2016estimating}, optimization methods \cite{newman2004finding, blondel2008fast} and methods based on dynamical processes \cite{pons2005computing, hastings2006community}.

Recently, geometric methods for community detection have gained increasing attention \cite{xue2017reliable, kovacs2021inherent, yang2021hidden}, with discrete Ricci curvatures playing a significant role.
Community detection methods utilizing discrete Ricci curvature can be broadly categorized into two approaches: incremental edge deletion methods and Ricci flow-based methods. A summary of existing curvature-based community detection methods is provided in Table~\ref{Tab:CommunityDetection}.

\setlength{\tabcolsep}{4pt}
\renewcommand{\arraystretch}{1.5}
\begin{table}[!ht]
\caption{\noindent \textbf{Applications of curvatures in community detection.}
Curvature-based community detection methods can be broadly categorized into incremental edge deletion and curvature flow-based approaches.
These methods are motivated by the observation that edges within communities typically exhibit positive curvature, while edges between communities often have negative curvature.
Incremental deletion methods iteratively remove negatively curved edges to reveal community structure, whereas curvature flow-based methods adjust edge weights using curvature flow and prune the graph using a cutoff rule.
The table provides details on the original publication where each method was proposed, the notion of curvature employed, and a summary of its primary contribution.}
\label{Tab:CommunityDetection}
\scriptsize
\centering
\hyphenpenalty=10000\exhyphenpenalty=10000
\begin{tabular}{>{\raggedright\arraybackslash}p{1.9cm} >{\raggedright\arraybackslash}p{1.9cm} >{\raggedright\arraybackslash}p{8.0cm}}
\toprule
\textbf{Reference} & \textbf{Curvature} & \textbf{Key Contribution} \\
\midrule

\multicolumn{3}{l}{\textit{Methods based on incremental edge deletion}} \\
\midrule
\cite{sia2019ollivier} & Ollivier & Demonstrates that networks with strong community structure exhibit bimodal curvature distributions. \\
\cite{gosztolai2021unfolding} & Ollivier & Proposes dynamical curvature via diffusion processes and geometric modularity for multiscale community detection. \\
\cite{fesser2024augmentations} & Forman & Extends Forman curvature to account for cycles of arbitrary lengths. \\

\midrule[1.5pt]
\multicolumn{3}{l}{\textit{Methods based on Ricci flow}} \\
\midrule
\cite{ni2019community} & Ollivier & Reveals communities by expanding inter-community edges and shrinking intra-community edges. \\
\cite{lai2022normalized} & Ollivier & Implements a normalized discrete flow for improved detection accuracy. \\
\cite{tian2023mixed} & Ollivier & Introduces a mixed-membership community detection algorithm for line graphs. \\

\bottomrule
\end{tabular}
\end{table}
\subsubsection{Curvature-based Community Detection}
Among the various notions of discrete Ricci curvature, Ollivier-Ricci curvature and Forman-Ricci curvature have been widely utilized in developing community detection algorithms \cite{sia2019ollivier, gosztolai2021unfolding, sia2022inferring, fesser2024augmentations}, with Ollivier-Ricci curvature being used in majority of these methods.
These algorithms are based on the following key observation. Edges that belong to dense subgraphs tend have positive curvature while those in sparse subgraphs have negative curvature.
In other words, edges connecting nodes \textit{within} a community generally have positive curvature, whereas edges connecting nodes \textit{between} communities often have negative curvature.
Notably, edges with highly negative curvature can be interpreted as \textit{bottlenecks} for information flow across the network.

Sia et al. \cite{sia2019ollivier}  observed that in stochastic block models as well as real-world networks with strong community structures, the distribution of the Ollivier-Ricci curvature of edges is bimodal with two distinct peaks.
The first peak corresponds to \textit{intra-community} edges with positive Ollivier-Ricci curvature, whereas the second peak corresponds to \textit{inter-community} edges with negative Ollivier-Ricci curvature.
They proposed a community detection algorithm that starts by incrementally removing negatively curved edges, and stops when no negatively curved edges exist. Once an edge is removed, the algorithm re-computes the Ollivier-Ricci curvature for the edges that are affected by its removal.

Gosztolai and Arnaudon \cite{gosztolai2021unfolding} developed a multiscale community detection algorithm using \textit{dynamical} Ollivier-Ricci curvature. Unlike the standard definition, where the probability mass is spread equally across the neighbors of a node, the dynamical approach incorporates a diffusion process around the node to define the probability masses used in the optimal transport problem. Subsequently, they introduced the concept of \textit{curvature gap} as an indicator of clusters in networks and proposed \textit{geometric modularity} to detect communities at multiple scales based on deviations from constant network curvature.

Augmentations of Forman-Ricci curvature were used by  Fesser et al. \cite{fesser2024augmentations} to develop a sequential edge deletion algorithm for community detection, similar to the approach proposed in \cite{sia2019ollivier}. These augmentations of Forman-Ricci curvature account for not only triangles but also cycles of arbitrary length within a graph. A significant advantage of using Forman-Ricci curvature and its augmentations over Ollivier-Ricci curvature for community detection is the reduced computational cost.

\subsubsection{Ricci flow-based Community Detection}
The key idea behind Ricci flow-based community detection methods involves modifying edge weights via Ricci flow and subsequently pruning them using a specified cutoff function.
In Riemannian manifolds, regions with high positive curvature are more densely packed than regions with negative curvature.
Under the process of Ricci flow \cite{hamilton1982three} regions with positive curvature shrink, while those with negative curvature expand, smoothing out any irregularities in the metric.
Since communities in networks correspond to positively curved regions in Riemannian manifolds, applying a discretized Ricci flow to networks can reveal underlying community structure.

Discrete Ricci flow based on Ollivier-Ricci curvture was first used for community detection by Ni et al. \cite{ni2019community}.
According to their algorithm, each edge is initialized with a weight which is updated through a discrete Ricci flow process, resulting in increased weights of negatively curved edges and reduced weights of positively curved edges. In other words, inter-community edges (typically negatively curved) are expanded, while intra-community edges (typically positively curved) are shrunk. Once the Ricci flow process concludes, a cutoff function, or surgery, is applied to remove edges with large weights, revealing the community structure. The choice of cutoff is arbitrary and influences the number of communities detected, with lower cutoff values producing more communities. This approach is especially useful for networks with hierarchical community structures, as varying the cutoff can uncover communities at different scales.

A normalized version of the discrete Ricci flow based community detection algorithm was proposed by Lai et al. \cite{lai2022normalized}.
Additionally, Tian et al. \cite{tian2023mixed} utilized Ollivier-Ricci curvature to develop a mixed-membership community detection algorithm based on the Ricci flow process on line graphs.

\subsection{Manifold Learning}
Manifold learning has become a significant research area of machine learning, pattern recognition and computer vision. It is is a type of dimensionality reduction technique that focuses on identifying and eliminating redundant dimensions in data by using the theory of manifolds \cite{tenenbaum2000global, roweis2000nonlinear}. 
Traditional manifold learning algorithms assume that the embedded manifold is globally or locally isometric to Euclidean space. However, one can relax this assumption by considering the curvature of the embedding manifold and its associated geometric flows, and effectively uncovering the non-Euclidean geometric structures of data distributions from high-dimensional complex spaces \cite{mahony2021representation}.
\setlength{\tabcolsep}{4pt}  
\renewcommand{\arraystretch}{1.5}      
\begin{table}[!ht]
\caption{\noindent \textbf{Applications of curvatures in manifold learning.}
Traditional manifold learning methods assume that the underlying data manifold is globally or locally isometric to Euclidean space. Curvature-based approaches relax this assumption and are applicable to data sampled from non-Euclidean manifolds.
These methods can be categorized into static methods, which directly incorporate curvature into the dimensionality reduction process, and dynamic methods based on Ricci flow.
The table provides details on the original publication where each method was proposed and a summary of its primary contribution.}
\label{Tab:ManifoldLearning}
\scriptsize
\centering
\hyphenpenalty=10000\exhyphenpenalty=10000
\begin{tabular}{>{\raggedright\arraybackslash}p{1.9cm} >{\raggedright\arraybackslash}p{10.4cm}}
\toprule
\textbf{Reference} & \textbf{Key Contribution} \\
\midrule

\multicolumn{2}{l}{\textit{Static methods based on direct inclusion of curvature information}} \\
\midrule
\cite{li2018curvature} & Proposes a curvature-based loss function for dimensionality reduction using local quadratic fitting. \\
\cite{fei2023nonlinear} & Computes Ollivier-Ricci curvature on affinity graphs derived from high-dimensional data. \\
\cite{saidi2025recovering} & Prunes spurious edges in nearest-neighbor graphs using Ollivier-Ricci curvature. \\

\midrule[1.5pt]
\multicolumn{2}{l}{\textit{Dynamic methods based on Ricci flow}} \\
\midrule
\cite{li2019applying} & Maps data features onto a sphere iteratively via Ricci flow. \\
\cite{li2022curvatureflow} & Introduces metric learning with a geometric potential linking Riemannian curvature to the metric. \\
\cite{zeng2010ricci} & Applies Ricci flow to 3D data points on 2D manifolds. \\
\cite{xu2010rectifying, xu2010regularising, xu2014ricci} & Corrects non-Euclidean pairwise dissimilarities using Ricci flow. \\
\cite{li2023geometry} & Develops a deep Riemannian metric learning framework that preserves geometric structure by regularizing local embeddings. \\
\bottomrule
\end{tabular}
\end{table}

Li \cite{li2018curvature} proposed curvature-aware manifold learning, which computes discrete Ricci curvature by local quadratic fitting of data points in ambient coordinates, and uses a reconstruction loss function based on the Ricci curvature of the high-dimensional data as a penalty to regularize the local similarity between any two points in each neighborhood.
Fei et al. \cite{fei2023nonlinear} introduced a nonlinear manifold learning approach that incorporates Ollivier-Ricci curvature to refine local similarity estimates. In particular, an affinity graph is constructed from high-dimensional data and Ollivier-Ricci curvature is computed on the edges of this affinity graph. A curvature-penalized affinity matrix is computed which reflects the embedded geometric structure of high-dimensional data. This curvature-penalized affinity matrix is combined with the traditional manifold learning algorithm to perform a generate low-dimensional vector representations.
Saidi et al. \cite{saidi2025recovering} proposed Ollivier–Ricci Curvature-based Manifold Learning and Recovery (ORC-MANL), an algorithm for pruning spurious edges in nearest-neighbor graphs.
When the data consists of noisy samples from a low-dimensional manifold, edges that shortcut through the ambient space exhibit more negative Ollivier–Ricci curvature than those lying along the intrinsic manifold. 
These candidate edges are filtered using graph distance thresholds to ensure that only genuine shortcuts are removed while preserving true manifold connections.
When used as a preprocessing step, ORC-MANL reconstructs the nearest-neighbor graph in a curvature-aware manner and yields more accurate low-dimensional embeddings.

There are also several manifold learning algorithms that utilize the notion of Ricci flow. For example, Li and Lu \cite{li2019applying} leverage Ricci flow to iteratively map data features onto a sphere. Under this method, the high-dimensional data is divided into overlapping patches, and subsequently, each patch is transformed into a spherical fragment using a discrete Ricci flow equation. These fragments are then aligned into a spherical subspace, and the spherical metric is employed to reduce the dimensionality.
Some Ricci-flow-based manifold learning methods avoid the assumption of positive curvature. For instance, a metric learning-based Ricci flow algorithm \cite{li2018curvature, li2022curvatureflow} introduces a geometric potential that connects Riemannian curvature with the metric. This approach employs an online learning model using Log-det divergence to iteratively evolve the metric towards curvature-energy balance.
Other nonlinear dimensionality reduction algorithms employ two dimensional discrete surface Ricci flow \cite{zeng2010ricci} (which is mainly applied to the three dimensional data points distributed on a two dimensional manifold), as well as using Ricci flow to rectify the pair-wise non-Euclidean dissimilarities among data points \cite{xu2010rectifying, xu2010regularising, xu2014ricci}.
Finally, Ricci flow has also been combined with deep learning to develop a deep Riemannian metric learning framework \cite{li2023geometry}, which not only regularizes the local neighborhoods connection of the embeddings at the hidden layer but also adapts the embeddings to preserve the geometric structure of the data.
A summary of these methods is provided in Table~\ref{Tab:ManifoldLearning}.

\subsection{Graph Neural Networks}
\setlength{\tabcolsep}{4pt}
\renewcommand{\arraystretch}{1.5}
\begin{table}[!ht]
\caption{\noindent \textbf{Applications of curvatures in graph neural networks (graph rewiring, structural and positional encoding).}
The table lists curvature-based methods that enhance the performance of GNNs through graph rewiring or structural and positional encoding.
Rewiring methods modify the connectivity of the graph by selectively removing or retaining edges based on curvature, thereby alleviating issues such as oversmoothing and oversquashing.
Structural and positional encoding methods enhance the expressivity of GNNs by assigning node or edge features with geometric information derived from curvature.
The table provides details on the original publication where each method was proposed, the name of the corresponding algorithm, the notion of curvature employed, and a summary of the primary contribution of each method.
}
\label{Tab:GNNApplications1}
\scriptsize
\centering
\hyphenpenalty=10000\exhyphenpenalty=10000
\begin{tabular}{>{\raggedright\arraybackslash}p{1.4cm} >{\raggedright\arraybackslash}p{2.5cm} >{\raggedright\arraybackslash}p{1.5cm} >{\raggedright\arraybackslash}p{6.3cm}}
\toprule
\textbf{Reference} & \textbf{Algorithm} & \textbf{Curvature} & \textbf{Key Contribution} \\
\midrule

\multicolumn{4}{l}{\textit{Graph Rewiring}} \\
\midrule
\cite{topping2021understanding} & Stochastic Discrete Ricci Flow (SDRF) & Forman & Strategically targets and removes negatively curved edges to mitigate oversquashing. \\
\cite{liu2023curvdrop} & CurvDrop & Ollivier & Strategic dropping of negatively curved edges at each GNN layer to improve information flow. \\
\cite{nguyen2023revisiting} & Batch Ollivier-Ricci Flow (BORF) & Ollivier & Addresses oversquashing/oversmoothing through selective removal of negatively/positively curved edges. \\
\cite{chen2024graph} & Global Discrete Ricci Flow & Ollivier & Combines Ricci flow with a spatial filtering mechanism, dynamically generating a topologically rich curvature flow mask matrix. \\
\cite{fesser2024mitigating} & AFR-k & Forman & Delivers scalable performance with linear time complexity for large-scale networks. \\

\midrule[1.5pt]
\multicolumn{4}{l}{\textit{Structural and Positional Encoding}} \\
\midrule
\cite{chen2023curvature} & Curvature-based Topology-aware Graph Embedding & Forman & Leverages Forman curvature computed on $k$-hop subgraphs to capture node-specific topological characteristics. \\
\cite{fesser2023effective} & Local Curvature Profile & Ollivier & Enriches node features with summary statistics of local curvature distributions. \\
\cite{lai2023deeper} & CurvPhormer & Ollivier & Employs curvature-based edge features to bias attention weights in graph transformers. \\

\bottomrule
\end{tabular}
\end{table}
\setlength{\tabcolsep}{4pt}
\renewcommand{\arraystretch}{1.5}
\begin{table}[!ht]
\caption{\noindent \textbf{Applications of curvatures in graph neural networks (feature aggregation, graph pooling).}
The table lists curvature-based methods that improve the performance of GNNs through feature aggregation or graph pooling.
Feature aggregation methods use local curvature information to guide the update of node representations during message passing.
Graph pooling methods use curvature to reduce the size of the graph while preserving essential structural information.
The table provides details on the original publication where each method was proposed, the name of the corresponding algorithm, the notion of curvature employed, and a summary of the primary contribution of each method.
}
\label{Tab:GNNApplications2}
\scriptsize
\centering
\hyphenpenalty=10000\exhyphenpenalty=10000
\begin{tabular}{>{\raggedright\arraybackslash}p{1.4cm} >{\raggedright\arraybackslash}p{2.6cm} >{\raggedright\arraybackslash}p{1.5cm} >{\raggedright\arraybackslash}p{6.3cm}}
\toprule
\textbf{Reference} & \textbf{Algorithm} & \textbf{Curvature} & \textbf{Key Contribution} \\
\midrule

\multicolumn{4}{l}{\textit{Feature Aggregation}} \\
\midrule
\cite{ye2019curvature} & Curvature Graph Network & Ollivier & Introduces learnable weight factors for feature aggregation as functions of Ollivier curvature. \\
\cite{li2022curvature} & Curvature Graph Neural Network & Ollivier & Implements predefined functional relationships between curvature values and weight factors. \\
\cite{wu2023curvagn} & CurvAGN & Forman & Combines edge curvatures with adaptive attention mechanisms in filtered graph representations. \\
\cite{shen2024curvature} & Curvature-Enhanced Graph Convolutional Network & Ollivier & Incorporates curvature into convolution weights to mitigate over-squashing effects. \\

\midrule[1.5pt]
\multicolumn{4}{l}{\textit{Graph Pooling}} \\
\midrule
\cite{sanders2023curvature} & CurvPool & Forman & Groups nodes connected by edges exhibiting similar curvatures. \\
\cite{feng2024graph} & ORC-Pool & Ollivier & Develops trainable pooling operators derived from Ollivier-based curvature flow. \\

\bottomrule
\end{tabular}
\end{table}

Graph neural networks (GNNs) provide a framework for extending deep learning to graph-structured data \cite{scarselli2008graph, defferrard2016convolutional, gilmer2017neural}.
Most GNNs operate on the message-passing mechanism, where vector messages are exchanged between neighboring nodes and updated using neural networks.
This mechanism enables GNNs to effectively utilize both node features and graph topology to learn rich node representations.
Prominent architectures like Graph Convolutional Networks (GCN) \cite{kipf2016semi}, GraphSAGE \cite{hamilton2017inductive}, Graph Attention Networks (GAT) \cite{velivckovic2017graph}, and Graph Isomorphism Networks (GIN) \cite{xu2018powerful} are specific instances of the message-passing framework and fall under the broader paradigm of \textit{geometric deep learning} \cite{bronstein2017geometric, bronstein2021geometric}. GNNs are widely applied to tasks such as link prediction, node classification, and graph classification.

Despite their success, GNNs face several limitations, which may arise either from undesirable properties of the input graph \cite{hamilton2017inductive, gasteiger2019diffusion} or fundamental constraints in the message-passing framework \cite{ye2019curvature}, often reducing their representational power. Discrete Ricci curvatures have proven effective in addressing these challenges.
Efforts to enhance GNN expressivity with discrete Ricci curvatures can be categorized into four approaches: \textit{graph rewiring} methods, \textit{structural and positional encoding} techniques, \textit{feature aggregation} based on local graph topology, and \textit{graph pooling}.
Table~\ref{Tab:GNNApplications1} lists the applications of discrete curvatures in graph rewiring and structural/positional encoding, and Table~\ref{Tab:GNNApplications2} lists their applications in feature aggregation and graph pooling.

\subsubsection{Graph Rewiring}
GNNs are designed to operate directly on input graphs, which can lead to significant downsides depending on the structure of the input graph. There are two key issues that impact their performance in applications: oversmoothing and oversquashing. Oversmoothing occurs when node features converge and become indistinguishable in deeper layers \cite{li2018deeper}, limiting the depth of GNNs and hindering their ability to capture complex relationships in the data. Oversquashing occurs when exponentially growing information is compressed into fixed-size node features, limiting the ability of GNNs to capture long-range dependencies \cite{alon2020bottleneck}. Graph rewiring, which involves modifying the set of edges within a graph, has been proposed as a preprocessing technique to mitigate the effects of oversmoothing and oversquashing \cite{karhadkar2022fosr, topping2021understanding, nguyen2023revisiting}.

Topping et al. \cite{topping2021understanding} were the first to study over-squashing in message-passing neural networks using discrete Ricci curvatures. Using a modified version on Forman-Ricci curvature, they showed that negatively curved edges create bottlenecks, and hence cause over-squashing.
In addition, they proposed Stochastic Discrete Ricci Flow, a graph rewiring method that mitigates over-squashing by surgically targeting negatively curved edges.
Nguyen et al. \cite{nguyen2023revisiting} introduced a unified theoretical framework using Ollivier-Ricci curvature to analyze oversmoothing and oversquashing in GNNs.
They demonstrated not only that negatively curved edges cause oversquashing, as previously observed, but also that positively curved edges lead to oversmoothing.
They proposed Batch Ollivier-Ricci Flow, a graph rewiring method designed to mitigate both the issues simultaneously.
Fesser and Weber \cite{fesser2024mitigating} proposed a scalable graph rewiring method based on augmentations of Forman-Ricci curvature. This approach has a linear-time complexity, making it well-suited for large-scale graphs.

Other notable graph machine learning methods utilizing curvature-based rewiring include CurvDrop \cite{liu2023curvdrop} and Global Discrete Ricci Flow (GDRF) \cite{chen2024graph}.
CurvDrop introduces a sampling layer based on Ollivier-Ricci curvature, dropping a portion of negatively curved edges at each GNN layer.
GDRF combines a global curvature flow algorithm based on Ollivier-Ricci curvature with a spatial filtering mechanism, dynamically generating a topologically rich curvature flow mask matrix, which is subsequently integrated into a graph transformer.

\subsubsection{Structural and Positional Encoding}
Structural encodings (SEs) and Positional encodings (PEs) provide GNNs with important structural information that it cannot learn on its own, but critical for enhancing their representational power \cite{dwivedi2021graph, dwivedi2023benchmarking}.
PE encode node positions within local substructures or global positions in the graph, often derived from distance metrics \cite{li2020distance} or spectral properties of the graph \cite{dwivedi2023benchmarking}.
SE capture structural similarity by embedding information about a node's subgraph or the global graph topology.
Both encodings can be designed to represent local or global information.
Examples of local SE include substructure counts \cite{zhao2021stars, bouritsas2022improving}, while global SE often encode graph-level statistics like diameter, girth, or connected components \cite{loukas2019graph}.
Discrete Ricci curvatures, which fundamentally capture the connectivity of local neighborhood of an edge, are particularly suited for local SE approaches.

Fesser and Weber \cite{fesser2023effective} introduced Local Curvature Profile (LCP), a structural node encoding using Ollivier-Ricci curvature, which is computed using summary statistics of the local curvature distribution.
They showed in a theoretical proof that LCP encodings are more expressive than the 1-WL test and message passing GNNs without encodings.
Chen et al. \cite{chen2023curvature} proposed Curvature-based Topology-Aware Graph Embedding (CTAGE) for graphs extracted from 3D point cloud data, with a particular focus on molecular representation learning. CTAGE utilizes k-hop discrete Ricci curvature to extract node structural encodings, integrating spatial structural information while maintaining the network's training complexity.
The k-hop discrete Ricci curvature is computed by generating a k-hop subgraph for a node and calculating the Forman curvature of the node within the subgraph.
Lai et al. \cite{lai2023deeper} introduced Curvphormer, a graph transformer that employs a degree-based SE for nodes and a curvature-based SE for the edges in the input graph. The node SE computed as the product of the node degree and a learnable vector, whereas the edge SE is computed by multiplying a limit-free definition of Ollivier-Ricci curvature with a learnable scalar.

\subsubsection{Message Aggregation}
Message aggregation, which refers to a specific set of rules for combining node features to update their representations, is a key component of the message passing framework.
Features are typically aggregated from the neighborhood of a node.
Most methods can be broadly divided into to two types.
The first set of methods aggregate features with equal importance, as seen in widely used models such as GIN \cite{xu2018powerful}, GraphSAGE \cite{hamilton2017inductive}, and Neural FPs \cite{duvenaud2015convolutional}.
The second set of methods aggregate features with different weights.
In GCNs, these weights depend on node degrees \cite{kipf2016semi}, whereas in GAT these weights are calculated using an attention mechanism \cite{velivckovic2017graph}.
Discrete Ricci curvatures are edge-based measures that capture the connectivity within the local neighborhoods of a pair of nodes.
Unlike node-based measures such as degree, discrete curvatures provide a more descriptive structural characterization. This unique property has been extensively utilized to improve traditional feature aggregation methods.

Ye et al. \cite{ye2019curvature} proposed the Curvature Graph Network (CurvGN), which leverages the local structural information captured by Ricci curvature to enhance feature aggregation in GCNs.
Specifically, they introduce a modified weight factor for feature aggregation which is defined as a learnable function of the Ollivier-Ricci curvature of an edge.
Li et al. \cite{li2022curvature} proposed the Curvature Graph Neural Network (CGNN), which also utilizes Ollivier-Ricci curvature to define the weight factor for feature aggregation.
However, unlike CurvGN which uses a learnable function to implicitly map edge curvature to weights, CGNN explicitly transforms Ricci curvature into neighboring node weights through the proposed Negative Curvature Transformation Module (NCTM) and Curvature Normalization Module (CNM).
Their method ensures that the relative magnitude of curvature is well preserved.
Shen et al. \cite{shen2024curvature} proposed the Curvature-Enhanced Graph Convolutional Network (CGCN), where they transform the Ollivier-Ricci curvature of an edge into a vector using an explicit function that is inversely related to the curvature, helping alleviate the over-squashing phenomenon.
These curvature-related vectors are then incorporated into the weight factors for graph convolution through a learnable function.
Finally, Wu et al. \cite{wu2023curvagn} introduced CurvAGN for graphs derived from 3D point cloud data, with a particular focus on protein-ligand binding affinity prediction.
The model utilizes Forman-Ricci curvature of edges in a sequence of filtered graphs to construct a curvature block, which is further integrated with an adaptive attention mechanism for feature aggregation.

\subsubsection{Graph Pooling}
Inspired by pooling layers in convolutional neural networks, recent advances in graph machine learning have introduced graph pooling operators to reduce graph size \cite{bianchi2020spectral, bianchi2020hierarchical, bianchi2023expressive}.
These operators aggregate nodes and their representations using certain criteria, producing a smaller graph with updated node representations. Graph pooling can improve memory efficiency and enhance the expressivity of GNNs.

Discrete Ricci curvatures have also been utilized to design efficient graph pooling operators.
Sanders et al. \cite{sanders2023curvature} proposed CurvPool, a pooling method based on a modified version of Forman-Ricci curvature. The key idea is to group nodes connected by edges with similar curvatures, merging dense clusters into single nodes and improving connectivity in sparsely connected regions.
CurvPool achieves linear runtime complexity with respect to the number of edges.
Feng and Weber \cite{feng2024graph} introduce ORC-Pool, a trainable pooling operator that uses Ollivier-Ricci curvature and an associated geometric flow to identify salient multi-scale structures in graphs.
ORC-Pool clusters nodes based on similarity in both graph topology and node attributes. The key distinction between the two pooling operators lies in their approach.
Specifically, CurvPool performs edge cuts guided by Ricci curvature, whereas ORC-Pool relies on curvature adjustment computed via Ricci flow.

\subsubsection{Graph Representation Learning}
\setlength{\tabcolsep}{4pt}
\renewcommand{\arraystretch}{1.5}
\begin{table}[!ht]
\caption{\noindent \textbf{Applications of curvatures in graph representation learning.}
Discrete curvatures have various applications in graph machine learning, extending beyond graph neural networks to the broader domain of graph representation learning.
The table provides details on the original publication where each method was proposed, the name of the corresponding algorithm, and a summary of the primary contribution of each method.}
\label{Tab:GRLApplications}
\scriptsize
\centering
\hyphenpenalty=10000\exhyphenpenalty=10000
\begin{tabular}{>{\raggedright\arraybackslash}p{1.4cm} >{\raggedright\arraybackslash}p{3.5cm} >{\raggedright\arraybackslash}p{7.1cm}}
\toprule
\textbf{Reference} & \textbf{Algorithm} & \textbf{Key Contribution} \\
\midrule
\cite{wang2021mixed} & Mixed-Curvature Multi-Relational GNN & Embeds entities and relations in a mixed-curvature product manifold for knowledge graph completion. \\
\cite{li2022curvatureGAN} & Curvature Graph Adversarial Network & Preserves local graph structure using curvature regularization on constant-curvature manifolds. \\
\cite{cho2023curve} & Fully Product-Stereographic Transformer & Generalizes Transformers to non-Euclidean spaces with learnable curvature. \\
\cite{sun2023deepricci} & DeepRicci & Refines graph structure and node features in latent Riemannian space to mitigate over-squashing. \\
\cite{yang2023kappahgcn} & Hyperbolic Curvature GNN & Embeds graphs in hyperbolic space using curvature-guided message passing. \\
\cite{zhang2023ricci} & Subgraph Episodic Memory & Sparsifies computation subgraphs via curvature for continual representation learning. \\
\cite{sun2024motif} & Motif-Aware Riemannian GNN & Utilizes a motif-aware Riemannian GNN with generative-contrastive learning on product manifolds of diverse curvature. \\
\cite{guo2025graphmore} & Graph Mixture of Riemannian Experts & Proposes a Riemannian mixture-of-experts model that routes nodes to curvature spaces via topology-aware gating. \\
\cite{hevapathige2025depth} & Curvature-Adaptive GNN & Integrates Bakry–\'Emery curvature through a learnable approximation, enabling adaptive message-passing depth. \\
\cite{liu2025a} & Mixed-Curvature Pretraining & Incorporates geometric pretraining for multi-task vehicle routing solvers via layer-wise subspace decomposition and curvature-informed fusion. \\
\bottomrule
\end{tabular}
\end{table}

Discrete Ricci curvatures have various applications in graph machine learning, extending beyond Graph Neural Networks to the broader domain of graph representation learning \cite{khoshraftar2024survey}.
Wang et al. \cite{wang2021mixed} introduced the Mixed-Curvature Multi-Relational Graph Neural Network (M$^{\mathrm{2}}$GNN) for knowledge graph completion, addressing the challenge of modeling heterogeneous relational structures in multi-relational graphs. 
The model embeds entities and relations in a mixed-curvature space constructed as a product manifold of Euclidean, spherical, and hyperbolic spaces. 
This formulation enables the embedding space to flexibly adapt to hierarchical and cyclic structures commonly found in real-world knowledge graphs. 
Li et al. \cite{li2022curvatureGAN} introduced Curvature Graph Generative Adversarial Network (CurvGAN).
In this model, global topology of the graph data is approximated by a Riemannian geometric space with constant curvature and local heterogeneous topology is characterized by Ollivier-Ricci curvature.

Cho et al. \cite{cho2023curve} proposed the Fully Product-Stereographic Transformer (FPS-T), generalizing Transformers to non-Euclidean spaces with learnable curvature. 
Each attention head in FPS-T operates within a stereographic model that unifies Euclidean, hyperbolic, and spherical geometries, enabling the network to learn embeddings and curvatures jointly in an end-to-end manner. 
Further, FPS-T was coupled with the Tokenized Graph Transformer, resulting in a kernelized non-Euclidean attention for graph representation learning.
Sun et al. \cite{sun2023deepricci} introduced DeepRicci, a self-supervised model that jointly refines graph structures using backward Ricci flow and enhances node features through geometric contrastive learning.
The Ricci flow process in DeepRicci is defined on a latent Riemannian space constructed using Ollivier-Ricci curvature of the edges in the input graph.
In this latent Riemannian space, different points can have varying curvatures.

Yang et al. \cite{yang2023kappahgcn} proposed the Hyperbolic Curvature Graph Neural Network (kHGCN), which also integrates global and local topology for learning a suitable representation of the graph.
However, unlike CurvGAN, kHGCN models the embedding space as a hyperbolic space with constant curvature, and incorporates Ollivier-Ricci curvature into the message-passing operator using hyperbolic curvature-aware propagation and a homophily constraint.
A key requirement for graph machine learning models is the ability to continually adapt to new tasks without compromising performance on previous ones.
To achieve this, Zhang et al. \cite{zhang2023ricci} proposed a Ricci curvature-based graph sparsification technique for continual graph representation learning.
They introduce Subgraph Episodic Memory (SEM) to store topological information as computation subgraphs.
These computation subgraphs are then sparsified using Forman-Ricci curvature to retain only the most informative nodes and edges.
A key advantage of their approach is reduced subgraph memory consumption while retaining key topological information about the graph.

Sun et al. \cite{sun2024motif} proposed the Motif-Aware Riemannian Graph Neural Network with Generative-Contrastive Learning (MotifRGC), a self-supervised framework designed to capture motif regularities within manifolds of diverse curvature. 
The model introduces a Diverse-Curvature Graph Convolutional Network that constructs a product manifold combining multiple curvature factors, supported by a numerically stable gyrovector kernel layer for mapping operations. 
Further, the model employs a Riemannian generative-contrastive scheme in which motif generators and discriminators operate on the manifold to learn node representations without supervision signals.
Guo et al. \cite{guo2025graphmore} introduced GraphMoRE, a Riemannian mixture-of-experts framework designed to mitigate topological heterogeneity in graph representation learning. 
Instead of embedding all nodes in a single constant-curvature or homogeneous product manifold, GraphMoRE employs a topology-aware gating mechanism that routes each node to the most suitable curvature space based on its local geometric properties. 
Multiple Riemannian experts operate in differentiated curvature spaces, including Euclidean, spherical, and hyperbolic spaces, thereby forming a heterogeneous manifold with point-wise varying curvature. 

Hevapathige et al. \cite{hevapathige2025depth} incorporated Bakry–\'Emery curvature into graph neural networks to jointly capture structural and diffusion-based aspects of information propagation. 
Their model introduces a learnable approximation of Bakry–\'Emery curvature that scales efficiently to large graphs, allowing curvature values to guide message-passing behavior dynamically. 
Specifically, an adaptive depth mechanism adjusts the number of layers per vertex based on its local curvature, where high-curvature regions undergo fewer propagation steps, while low-curvature regions benefit from deeper layers.
Finally, Liu et al. \cite{liu2025a} incorporated curvature-aware geometric representation learning into combinatorial optimization by proposing a mixed-curvature pretraining framework for multi-task vehicle routing solvers.
Their framework decomposes the feature space of each encoder layer into several subspaces embedded in Euclidean, hyperbolic, or spherical geometries, which are subsequently combined through a curvature-informed fusion mechanism.

All the above approaches differ from traditional embedding methods, which typically assume a representation space with uniform curvature, such as Euclidean, hyperbolic, or spherical spaces.
A summary of these approaches is provided in Table~\ref{Tab:GRLApplications}.

\section{Conclusions}\label{sec8}
This article surveyed  discrete curvature models and their practical applications in data analysis and learning.
In particular, we consolidated existing discrete curvature models and introduced a step-by-step computational pipeline for extracting geometric information from data. To make discrete curvatures more accessible to machine learning community, we included several figures that illustrate their computation and the intuition behind their formulations.
Additionally, we discussed important concepts associated with curvature such as Ricci flow on discrete spaces, and reviewed a diffusion-based curvature formulation applicable to point clouds sampled from manifolds embedded in high-dimensional Euclidean spaces.
Finally, we reviewed applications of discrete curvatures in both supervised and unsupervised machine learning tasks, including graph representation learning, community detection, and manifold learning.

There remain many challenges and opportunities in the development of discrete curvatures.
A key direction is to develop a theoretical framework that classifies and relates the various notions of discrete curvature. One promising approach is the semigroup characterization \cite{jost2021characterizations}, which has been shown to form the basis of some of the existing definitions.
There is a also clear need for systematic empirical comparisons of different notions of discrete curvature.
Previous studies on both model and real-world network datasets have reported strong correlations among Forman-Ricci, Ollivier-Ricci, and Bakry-\'Emery curvatures \cite{samal_comparative_2018, mondal2024bakry}.
In our own demonstration on molecular structures, we similarly observed areas of both agreement and divergence in the qualitative behavior of different curvature models.
These findings underscore the need for standardized benchmarks to enable broader evaluations and draw robust conclusions about the behavior and geometric insights offered by each curvature model. Another important direction is the extension of curvature models beyond simple graphs to more complex data representations, such as hypergraphs and simplicial complexes.
Recent work has generalized Forman-Ricci and Ollivier-Ricci curvature to hypergraphs and introduced corresponding notions of Ricci flow \cite{garcia2025finding, hacquard2024hypergraph}.
Finally, we identify future directions for applications in machine learning.
Most current methods rely on Forman-Ricci or Ollivier-Ricci curvature, however incorporating other definitions could offer a diverse set of geometric machine learning models. Notably, Bakry-\'Emery curvature has recently been integrated into graph neural networks \cite{hevapathige2025depth}. Additionally, there is a need to characterize higher-order structures such as simplices, which are central to the emerging topological deep learning (TDL) framework \cite{zia2024topological}.
Notions of discrete curvature, such as Forman curvature, are well-suited for this as they naturally extend to higher-dimensional simplices.

To summarize, the study of discrete curvatures has become increasingly popular, along with their applications to real-world datasets and machine learning.
We anticipate that more definitions of discrete curvature will continue to emerge, alongside powerful geometric machine learning algorithms that utilize curvature.
By offering practitioners and researchers a guide to current formulations of discrete curvature, their computation on data, and state-of-the-art algorithms, this work aims to support future developments in the field.

\appendix
\section{Additional discussion on sectional curvature of general metric spaces}\label{secA1}
For three points $x_1, x_2, x_3$ in a metric space $(X,d)$, sectional curvature is defined as follows.
$$
\rho(x_1, x_2, x_3) := \inf_{x \in X} \max_{i=1,2,3} \frac{d(x_i, x)}{r_i}
$$
For a given point $x \in X$, we consider the largest ratio of its distance from the three points $x_1$, $x_2$, $x_3$ and the radii of the respective closed balls $r_1, r_2$ and $r_3$ respectively. This operation is equivalent to expanding the radii of the three closed balls by a common scaling factor while ensuring that the point $x$ is included in their common intersection. Finally, we consider only those points $x \in X$ for which the radii need to be scaled the least in order to include them in the common intersection. The smallest value of this scaling factor corresponds to the sectional curvature of the three points $x_1$, $x_2$ and $x_3$.

For complete metric spaces $1 \leq \rho(x_1, x_2, x_3) \leq 2$. The lower bound is achieved for any three points in a tripod space. In other words, for tripod spaces, a pairwise intersection of three closed balls guarantees a common intersection. Further, if the closed balls intersect pairwise at their boundaries, the common intersection results in a point $m \in X$ which is the median. As a result, sectional curvature is an invariant associated with triples of points in a metric space which measures the deviation from tripod property.  Larger values of sectional curvature indicate that a metric space is further away from being a tripod space.

\subsection{Extension to more than three points}
Note that tripod spaces are a special case of hyperconvex spaces.
A geodesic metric space $(X, d)$ is hyperconvex if every family of pairwise intersecting closed balls $\{B(x_i, r_i)\}_{i \in I}$ has a non-empty intersection.
If we treat hyperconvex spaces as model spaces, it is possible to extend the definition of sectional curvature to arbitrary number of points.
Specifically, for a family $\{x_i\}_{i \in I}$ of points in $X$ and positive real numbers $\{r_i\}_{i \in I}$, one can assign the number
$$
\rho(\{x_i\}, \{r_i\}) := \inf_{x \in X} \sup_{i \in I} \frac{d(x_i, x)}{r_i}. \nonumber
$$
Subsequently, one can define the expansion constant $\rho(\{x_i\})$ associated with the family of points $\{x_i\}_{i \in I}$ as the supremum of these numbers over all positive real numbers $\{r_i\}_{i \in I}$ such that $r_i + r_j \geq d(x_i, x_j)$ for $i, j \in I$.

Thus, the expansion constant for an arbitrary number of points measures the deviation of a space from hyperconvexity. Note that the expansion constant of hyperconvex spaces is equal to $1$.  We refer the reader to \cite{joharinad_topology_2019, joharinad_geometry_2022} for a detailed discussion on the relationship between sectional curvature and hyperconvexity.

\subsection{Upper and lower bounds on graphs}
For complete metric spaces, the sectional curvature satisfies $1 \leq \rho(x_1, x_2, x_3) \leq 2$, where the lower bound is achieved in tripod spaces (and hyperconvex spaces in general).
Notably, these bounds also hold true for graphs, providing useful insights about their connectivity.
Among the previously discussed examples, star graphs, spider graphs, and path graphs satisfy the lower bound of $1$, while triangular graphs satisfy the upper bound of $2$.
More generally, \textit{the lower bound holds for trees, while the upper bound holds for complete graphs}.
An intuitive explanation for these general bounds is as follows.
Star graphs, spider graphs, and path graphs naturally arise in tree-like structures, whereas any three vertices in a complete graph form a triangle.
Consequently, trees and complete graphs serve as "model" structures that allow for comparisons with other graph topologies.
Remarkably, the sectional curvature of a graph captures its deviation from being either a tree or a fully connected graph.
In other words, sectional curvature provides a quantitative measure of where a graph lies on the spectrum between tree-like and fully connected structures.

\bibliographystyle{siamplain}
\bibliography{references}
\end{document}